\documentclass[11pt]{article}

\usepackage{acl}
\usepackage{times}
\usepackage{latexsym}
\usepackage{graphicx}
\usepackage{rotating}
\usepackage{forest}
\usetikzlibrary{arrows.meta,calc,shapes.misc,backgrounds}
\usepackage{fontawesome5}
\usepackage{siunitx}
\usepackage{amssymb}
\usepackage[T1]{fontenc}
\usepackage[utf8]{inputenc}
\usepackage{microtype}
\usepackage{inconsolata}
\usepackage{float}
\usepackage{placeins}
\usepackage{dblfloatfix}
\usepackage{booktabs}
\usepackage{multirow}
\usepackage{array}
\usepackage{xcolor}
\usepackage{colortbl}
\usepackage[normalem]{ulem}
\DeclareUnicodeCharacter{202F}{\,}
\DeclareUnicodeCharacter{2212}{-}
\usepackage{amsmath,amssymb,amsfonts,bm}

\DeclareSIUnit{\pp}{pp}

\usepackage{hyperref}
\usepackage{url}

\usepackage{pifont}           
\usepackage{tikz}             
\usetikzlibrary{positioning,fit,backgrounds}
\usepackage{enumitem}         
\usepackage[many,breakable]{tcolorbox}   

\newcommand{\cmark}{\ding{51}}
\newcommand{\xmark}{\ding{55}}

\definecolor{VioletB}{HTML}{7F77DD}
\definecolor{TealB}  {HTML}{1D9E75}
\definecolor{TealD}  {HTML}{0F6E56}
\definecolor{TealF}  {HTML}{E1F5EE}
\definecolor{AmberB} {HTML}{BA7517}
\definecolor{AmberF} {HTML}{FAEEDA}
\definecolor{CoralB} {HTML}{D85A30}
\definecolor{CoralF} {HTML}{FAECE7}
\definecolor{RedB}   {HTML}{E24B4A}
\definecolor{RedF}   {HTML}{FCEBEB}
\definecolor{GreenB} {HTML}{639922}
\definecolor{GreenF} {HTML}{EAF3DE}
\definecolor{GrayB}  {HTML}{888780}
\definecolor{GrayF}  {HTML}{F1EFE8}
\definecolor{GrayD}  {HTML}{3d3d3a}
\definecolor{RuleC}  {HTML}{CCCBC5}
\definecolor{Txt2}   {HTML}{555552}
\definecolor{Txt3}   {HTML}{999994}

\definecolor{RowV}   {HTML}{EEEDFE}
\definecolor{RowT}   {HTML}{E1F5EE}
\definecolor{RowA}   {HTML}{FAEEDA}
\definecolor{RowC}   {HTML}{FAECE7}
\definecolor{RowR}   {HTML}{FCEBEB}
\definecolor{RowG}   {HTML}{EAF3DE}

\newcommand{\CV}[1]{\cellcolor{RowV}#1}
\newcommand{\CT}[1]{\cellcolor{RowT}#1}
\newcommand{\CA}[1]{\cellcolor{RowA}#1}
\newcommand{\CCo}[1]{\cellcolor{RowC}#1}
\newcommand{\CR}[1]{\cellcolor{RowR}#1}
\newcommand{\CG}[1]{\cellcolor{RowG}#1}

\title{\menalogo\hspace{0.1cm}EiCAP: Beyond Fluency, Probing and Improving Emotional Intelligence in LLMs via Psychologically Grounded Multi-Turn Dialogue}

\newcommand\menalogo{\raisebox{-4pt}[0cm][0cm]{\includegraphics[width=2.2em]{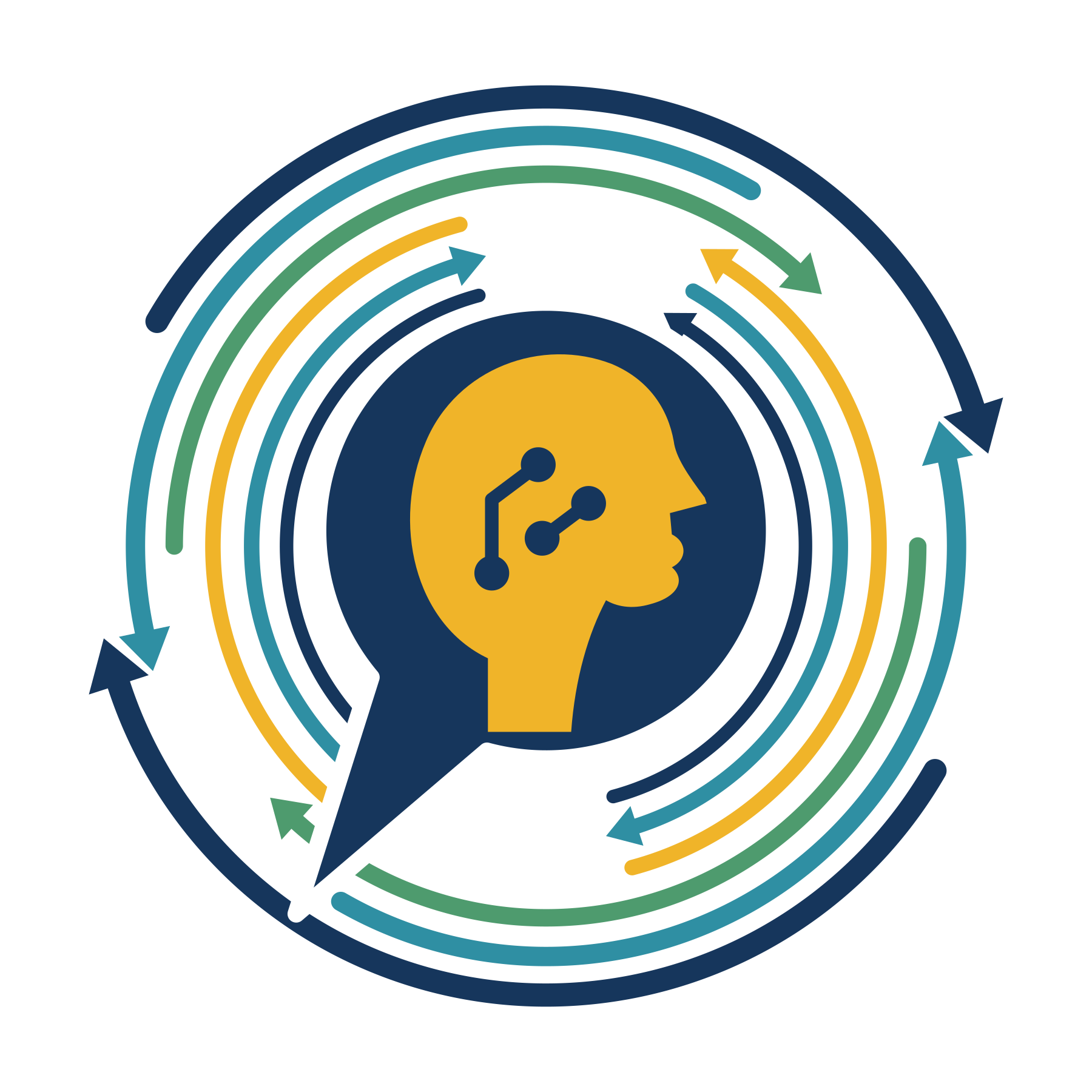}}}
\definecolor{logoRed}{HTML}{B22222}

\author{
\textbf{Nizi Nazar}$^{*\dagger}$ \quad
\textbf{Pardis Sadat Zahraei}$^{*\ddagger}$ \quad
\textbf{Dilek Hakkani-Tür}$^{\ddagger}$ \\[0.5em]
\textbf{Natasa Milic-Frayling}$^{\dagger}$ \quad
\textbf{Ehsaneddin Asgari}$^{c,\dagger}$ \\[1em]
$^{\dagger}$Qatar Computing Research Institute (QCRI), Hamad Bin Khalifa University \\
$^{\ddagger}$University of Illinois Urbana-Champaign (UIUC) \\
\texttt{\{nnazar,nmilicfrayling,$^{c}$\uline{easgari}\}@hbku.edu.qa \quad \{zahraei2,dilek\}@illinois.edu}
}

\begin{document}
\maketitle

\begin{abstract}
Large Language Models increasingly serve in emotionally sensitive roles (mental health support, education, and crisis response), yet they lack a principled framework for assessing
or improving Emotional Intelligence~(EI).
We introduce \textbf{EiCAP}, a unified, psychologically grounded \textbf{six-layer EI
taxonomy} operationalized into two complementary resources.
\textbf{EiCAP-Bench} is a multi-turn, one-vs.-three forced-choice evaluation suite
(3,174 probes across 24 subcategories) with cross-turn dependencies that reflect real
conversational EI demands.
\textbf{EiCAP-SFT} is a 152,820-dialogue supervision corpus aligned to the same
taxonomy, enabling controlled, interpretable fine-tuning.
Two key findings emerge:
(1)~\emph{Generic conversational SFT does not confer EI}: fine-tuning on UltraChat 
yields no significant gain in any of 24 subcategories (macro 24.6\%, near chance 25\%).
(2)~Applying EI-grounded LoRA (${\sim}0.8\%$ parameters) \emph{directly} to
Qwen-2.5-7B-Base achieves \textbf{24/24 subcategories significant}, macro
$\mathbf{75.33\%}$ ($+51.7$~pp over Base,  $+37.1$~pp over Instruct).
Crucially, an ablation shows the UltraChat pre-stage is \emph{counterproductive}
($-21.4$~pp): direct EI-grounded training is both necessary and sufficient\footnote{All benchmarks, corpora, and models will be publicly released upon acceptance.  \\ $^{*}$Equal contribution.\\$^{c}$Corresponding author}.

\end{abstract}

\section{Introduction}
\label{sec:intro}

Consider a user who writes: \textit{``Part of me thinks I should quit this job because
it drains me, but I keep wondering if I'm just being dramatic. I don't trust my own
judgment anymore.''} A model fine-tuned on generic dialogue responds with cascading
self-doubt (\textit{``I'm not sure if I'm just being lazy\ldots{} I'm not sure if I'm
just not trying hard enough''}), repeated many times. A model fine-tuned with
EiCAP-SFT, by contrast, grounds its reply in the specific emotional tension, validates
the ambivalence, and gently asks a clarifying question without amplifying the distress.
This difference between dialog fluency and genuine emotional intelligence is what
EiCAP is designed to measure and improve.
LLMs are deployed in emotionally sensitive contexts (mental health support, crisis
response, education) where misreading affect causes real harm~\cite{Fitzpatrick2017,
ou2023dialogbench}. Yet the field still lacks a principled framework that jointly
\emph{defines} EI for LLMs, \emph{measures} it rigorously, and \emph{provides matched
training data} for systematic improvement~\cite{Sabour2024EmoBench}.

\paragraph{The missing piece: a grounded taxonomy.}
Prior EI benchmarks aggregate coarse task-level scores without grounding in a
shared psychological theory or a systematic coverage of the EI space. This creates
two practical problems. First, we cannot tell \emph{which} EI skills are deficient:
a model that fails ``emotion recognition'' in one benchmark may be failing
basic-label identification, mixed-affect detection, causal appraisal, or Theory of Mind; all lumped together. Second, we cannot design targeted training because we do not know
what concepts to supervise. \emph{A principled taxonomy that maps psychological theory
to LLM behaviors is the missing architectural piece on which both evaluation and
improvement depend.}

\paragraph{Our approach: EiCAP.}
We introduce \textbf{EiCAP}, addressing this gap in four stages.
(i)~We derive a \textbf{six-layer EI taxonomy} grounded in Ekman, Plutchik, Russell,
Goleman, and Lazarus and operationalized into 24 subcategories and 53 leaf nodes covering
basic emotion recognition, mixed affect, causal appraisal, distress escalation, Theory
of Mind, and culturally sensitive response.
(ii)~We build \textbf{EiCAP-Bench} from the taxonomy: 3,174 multi-turn forced-choice
probes with cross-turn dependencies.
(iii)~We build \textbf{EiCAP-SFT}: 152,820 taxonomy-aligned dialogues for controlled
fine-tuning.
(iv)~We run controlled experiments that isolate what does and does not confer EI skill.

\paragraph{Contributions.}
\begin{enumerate}[leftmargin=*,topsep=2pt,itemsep=1pt]
  \item \textbf{EiCAP taxonomy:} a six-layer psychologically grounded EI framework
    with 24 subcategories, each mapped to a specific psychological theory and normative
    LLM response principle. First taxonomy designed for joint benchmark and training
    corpus construction.
  \item \textbf{EiCAP-Bench:} 3,174 multi-turn one-vs.-three probes with cross-turn
    dependency, covering all 24 subcategories. Validated against a held-out
    cross-generator probe set ($\kappa{=}0.976$).
  \item \textbf{EiCAP-SFT:} 152,820-dialogue taxonomy-aligned training corpus.
    First matched SFT resource for comprehensive multi-layer EI fine-tuning.
\end{enumerate}

\section{Related Work}
\label{sec:related}

\paragraph{EI benchmarks and taxonomies.}
EmoBench~\cite{Sabour2024EmoBench} evaluates emotion recognition and appraisal via
single-turn MCQ over short vignettes, building on the Mayer--Salovey ability model.
EIBENCH~\cite{zhao2024both} covers recognition, causal reasoning, and empathetic
response. EmotionQueen~\cite{qi2025emoassist} targets four event-recognition tasks.
SOCIAL-IQA~\cite{sap2019socialiqacommonsensereasoningsocial} covers social inference
but is single-turn. As Table~\ref{tab:benchmark_comparison} shows, \emph{none} provides
multi-turn evaluation with cross-turn dependency, a full EI taxonomy with 24
subcategories, or a matched SFT corpus. EiCAP-Bench is 4.6$\times$ harder than
EmoBench on discriminability: Qwen-2.5-7B-Instruct scores 87\% on EmoBench but only
38\% on EiCAP-Bench (\S\ref{app:difficulty}).

\paragraph{Empathetic dialogue.}
EmpatheticDialogues~\cite{rashkin2019empatheticopendomainconversationmodels} is
multi-turn but lacks taxonomy annotations or normative response constraints.
ESConv~\cite{liu-etal-2021-towards} targets counseling strategies within a narrow domain.
EmoDynamiX~\cite{Wan_2025} models emotion dynamics on ESConv.
We show empirically that fine-tuning on either corpus \emph{catastrophically degrades}
EI performance on EiCAP-Bench (\S\ref{sec:results}, Table~\ref{tab:ablation})
.

\paragraph{Emotional fine-tuning.}
SoulChat~\cite{chen2023soulchat} and Emotional CoT~\cite{Li2023ECoT} show that
domain-specific empathy data helps in narrow settings but do not provide a comprehensive
taxonomy-grounded evaluation or corpus. Our central experimental finding extends this
insight: improvement requires data organized by a principled, multi-dimensional taxonomy
covering the full EI space; generic empathy is not enough.

\section{The EiCAP Taxonomy}
\label{sec:tax}

\begin{figure*}[p]
\centering
\resizebox{\textwidth}{!}{
\forestset{
  forked edges/.style={
    for tree={
      edge path={
        \noexpand\begin{scope}[on background layer]
          \noexpand\path [\forestoption{edge}]
            (!u.parent anchor) -- (.child anchor)
            \forestoption{edge label};
        \noexpand\end{scope}
      },
    },
  },
}
\begin{forest}
  forked edges,
  for tree={
    child anchor=west, parent anchor=east,
    grow'=east, anchor=west, base=left,
    font=\small, rectangle, draw=black!60, rounded corners,
    minimum height=1.8em, minimum width=3em,
    edge+={black!50, line width=0.8pt},
    s sep=2pt, l sep=1.5em,
    inner xsep=0.35em, inner ysep=0.45em,
    line width=0.7pt, text width=8.5em,
    where level=1{text width=5em, font=\small}{},
    where level=2{text width=8.5em}{},
    where level=3{text width=6.5em}{},
    where level=1{no edge}{},
    ver/.style={rotate=90, child anchor=north, parent anchor=south,
      anchor=center, text width=7em, font=\normalsize\bfseries, fill=gray!15},
    leaf/.style={text opacity=1, fill opacity=1, text=black,
      text width=41em, font=\small, inner xsep=0.4em, inner ysep=0.5em, draw},
    L1/.style={fill=violet!30, draw=violet!60, text=black, font=\small\bfseries},
    L1a/.style={fill=violet!18, draw=violet!45},
    L1b/.style={fill=violet!9,  draw=violet!30},
    L1leaf/.style={leaf, fill=violet!9},
    L2/.style={fill=teal!30,   draw=teal!60,   text=black, font=\small\bfseries},
    L2a/.style={fill=teal!18,  draw=teal!45},
    L2b/.style={fill=teal!9,   draw=teal!30},
    L2leaf/.style={leaf, fill=teal!9},
    L3/.style={fill=orange!35, draw=orange!65, text=black, font=\small\bfseries},
    L3a/.style={fill=orange!20, draw=orange!50},
    L3b/.style={fill=orange!10, draw=orange!32},
    L3leaf/.style={leaf, fill=orange!10},
    L4/.style={fill=cyan!32,   draw=cyan!62,   text=black, font=\small\bfseries},
    L4a/.style={fill=cyan!18,  draw=cyan!45},
    L4b/.style={fill=cyan!9,   draw=cyan!30},
    L4leaf/.style={leaf, fill=cyan!9},
    L5/.style={fill=red!28,    draw=red!55,    text=black, font=\small\bfseries},
    L5a/.style={fill=red!16,   draw=red!42},
    L5b/.style={fill=red!8,    draw=red!28},
    L5leaf/.style={leaf, fill=red!8},
    L6/.style={fill=green!32,  draw=green!62,  text=black, font=\small\bfseries},
    L6a/.style={fill=green!18, draw=green!45},
    L6b/.style={fill=green!9,  draw=green!30},
    L6leaf/.style={leaf, fill=green!9},
  },
  [EiCAP Taxonomy, ver
    [Foundation {(Receiving)}, L1
      [Basic Emotion Labels~\cite{ekman1992,russell1980}, L1a
        [Happiness \& Valenced Basics~\cite{fredrickson2001}, L1b
          [Happiness~\cite{fredrickson2001}{,} Sadness~\cite{rottenberg2003}{,}
           Anticipatory Fear~\cite{barlow2002}{,} Immediate Fear~\cite{panksepp1998}{,}
           Anger~\cite{averill1983}{,} Positive Surprise~\cite{ekman2003}{,}
           Negative Surprise~\cite{Lazarus1991EmotionAdaptation}{,}
           Relief~\cite{fredrickson2001}{,} Disgust~\cite{rozin1999individual}, L1leaf]
        ]
      ]
      [Confidence \& Uncertainty~\cite{tversky1974}, L1a
        [Detection Confidence~\cite{ekman2003}, L1b
          [High Confidence~\cite{ekman2003}{,} Low Confidence~\cite{tversky1974}{,}
           Overlapping Cues~\cite{barrett2006}{,}
           Probability-Based Calibration~\cite{kahneman2011}{,}
           Explainability \& Reasoning Traces~\cite{miller2019}, L1leaf]
        ]
      ]
      [Contextual Lexical Cues~\cite{pennebaker2001}, L1a
        [Lexical Cue Types~\cite{fellbaum1998}, L1b
          [Emotion-Specific Words~\cite{fellbaum1998}{,} Amplifiers~\cite{wiebe2005}{,}
           Negative Lexical Cues~\cite{turney2003}{,}
           Idiomatic Phrases~\cite{kreuz1995}, L1leaf]
        ]
      ]
      [Neutrals vs.\ Emotional~\cite{schwarz2012}, L1a
        [Content Types~\cite{grice1975}, L1b
          [Neutral Statements~\cite{grice1975}{,}
           Emotional Statements~\cite{schwarz2012}{,}
           Blended Content~\cite{wiebe2005}, L1leaf]
        ]
      ]
      [Edge Cases~\cite{kreuz1995}, L1a
        [Ambiguous~\cite{wiebe2005}, L1b
          [Ambiguous Statements~\cite{wiebe2005}, L1leaf]
        ]
        [Sarcasm or Irony~\cite{kreuz1995}, L1b
          [Mild Sarcasm~\cite{attardo2000}{,} Heavy Sarcasm~\cite{gibbs2000}, L1leaf]
        ]
        [Mixed Cues~\cite{plutchik2001}, L1b
          [Mixed or Conflicting Cues~\cite{plutchik2001}, L1leaf]
        ]
      ]
      [Acknowledgment Types~\cite{clark1996}, L1a
        [Response Styles~\cite{fredrickson2001}, L1b
          [Positive Reinforcement~\cite{fredrickson2001}{,}
           Sympathetic/Supportive~\cite{averill1983}{,}
           Neutral/Clarifying~\cite{clark1996}, L1leaf]
        ]
      ]
      [Politeness \& Style~\cite{brown1987}, L1a
        [Appropriate Politeness \& Style~\cite{brown1987}, L1leaf]
      ]
    ]
    [Foundation {(Responding)}, L2
      [Minimal Emotional Inference~\cite{russell2003}, L2a
        [Minimal Emotional Inference~\cite{russell2003}, L2leaf]
      ]
      [Error Recovery~\cite{brennan1998}, L2a
        [Error Recovery~\cite{brennan1998}, L2leaf]
      ]
    ]
    [Dimensional\\~\cite{russell1980}, L3
      [Valence Scale~\cite{bradley1999}, L3a
        [Valence Scale~\cite{bradley1999}, L3leaf]
      ]
      [Arousal Scale~\cite{lang1993}, L3a
        [Arousal Scale~\cite{lang1993}, L3leaf]
      ]
      [Mixed Emotion Detection~\cite{plutchik2001}, L3a
        [Mixed Emotion Detection~\cite{plutchik2001}, L3leaf]
      ]
      [Temporal Emotion~\cite{izard1991}, L3a
        [Emotion Trajectories~\cite{fredrickson2001}, L3b
          [Improving~\cite{fredrickson2001}{,} Worsening~\cite{rottenberg2003}{,}
           Fluctuating~\cite{barrett2006}, L3leaf]
        ]
      ]
    ]
    [Appraisal\\~\cite{Lazarus1991EmotionAdaptation}, L4
      [Cause/Consequence Inference~\cite{tversky1974}, L4a
        [Causal Attribution~\cite{Lazarus1991EmotionAdaptation}, L4b
          [External Causes~\cite{Lazarus1991EmotionAdaptation}{,}
           Internal Causes~\cite{barrett2006}{,}
           Implicit Causes~\cite{wilson2002}, L4leaf]
        ]
      ]
      [Prioritization of Response~\cite{tversky1974}, L4a
        [Prioritization of Response~\cite{tversky1974}, L4leaf]
      ]
      [Cultural Sensitivity~\cite{mesquita2010}, L4a
        [Cultural Sensitivity~\cite{mesquita2010}, L4leaf]
      ]
      [Two-Turn Context Tracking~\cite{kahneman2011}, L4a
        [Two-Turn Context Tracking~\cite{kahneman2011}, L4leaf]
      ]
      [Ethical Safeguards~\cite{bostrom2014}, L4a
        [Distress Escalation Detection~\cite{mitchell2021}, L4b
          [Distress Escalation Detection~\cite{mitchell2021}, L4leaf]
        ]
      ]
    ]
    [Values\\~\cite{bostrom2014}, L5
      [Intervention \& Escalation~\cite{mitchell2021}, L5a
        [Distress Tiers~\cite{rozin1999individual}, L5b
          [Tier~1: Mild Distress~\cite{rozin1999individual}{,}
           Tier~2: Moderate/Serious Distress~\cite{bostrom2014}{,}
           Tier~3: Crisis (Severe Distress)~\cite{mitchell2021}, L5leaf]
        ]
      ]
      [Long-Term Emotional Memory~\cite{kahneman2011}, L5a
        [Long-Term Emotional Memory~\cite{kahneman2011}, L5leaf]
      ]
    ]
    [Social\\~\cite{mesquita2010}, L6
      [Theory of Mind~\cite{premack1978}, L6a
        [Theory of Mind~\cite{premack1978}, L6leaf]
      ]
      [Conflict Mediation~\cite{brown1987}, L6a
        [Conflict Types~\cite{tomasello2005}, L6b
          [Personal Misunderstandings~\cite{tomasello2005}{,}
           Deep-Seated Conflicts~\cite{mesquita2010}, L6leaf]
        ]
      ]
      [Cultural Sensitivity~\cite{hofstede2001}, L6a
        [Cultural Dimensions~\cite{mesquita2010}, L6b
          [Regional/Generational Differences~\cite{mesquita2010}{,}
           Polite Navigation/Taboo Topics~\cite{brown1987}, L6leaf]
        ]
      ]
      [Social Ethics \& Sensitivity~\cite{bostrom2014}, L6a
        [Social Ethics \& Sensitivity~\cite{bostrom2014}, L6leaf]
      ]
    ]
  ]
\end{forest}
}
\caption{\textbf{EiCAP Taxonomy.}
Six-layer hierarchy grounded in established psychological theories,
operationalized into 24 subcategories and 53 leaf nodes.
Each layer assigns a \emph{distinct representational role} to an existing psychological
framework:
\textcolor{VioletB}{\textbf{Foundation (Receiving)}} (violet): Ekman's basic emotions
and Plutchik's wheel provide the \emph{label space} for discrete affect categories;
\textcolor{TealB}{\textbf{Foundation (Responding)}} (teal): Russell's
minimal-inference framework specifies \emph{reply formulation under uncertainty};
\textcolor{AmberB}{\textbf{Dimensional}} (orange): Russell's circumplex and Plutchik's
wheel provide the \emph{geometric valence$\times$arousal structure} for continuous and
mixed affect;
\textcolor{CoralB}{\textbf{Appraisal}} (cyan): Lazarus's appraisal theory supplies
the \emph{causal inference rules} linking events to affect;
\textcolor{RedB}{\textbf{Values}} (red): Goleman's framework supplies \emph{normative
response constraints} (tiered escalation, persistence, long-term memory);
\textcolor{GreenB}{\textbf{Social}} (green): Theory of Mind, conflict mediation, and
cultural sensitivity cover \emph{multi-agent and cross-cultural EI}.
The taxonomy is the architectural backbone from which EiCAP-Bench probes and EiCAP-SFT
dialogues are both derived.}
\label{fig:taxonomy}
\end{figure*}

\subsection{Layer Descriptions}
\label{sec:tax_desc}

The taxonomy is structured in six progressively abstract layers.
\emph{We do not propose new psychological theory}: our contribution is
operationalizing four well-validated frameworks into a unified, LLM-actionable structure
with matched evaluation and training resources.

\noindent\textbf{Layer 1: Foundation (Receiving)} covers how an agent
\emph{perceives} emotional signals: 7 subcategories spanning basic emotion labels,
confidence/uncertainty handling, contextual lexical cues, neutral vs.\ emotional
content, edge cases (sarcasm, ambiguity, mixed cues), acknowledgment types, and
appropriate politeness.

\noindent\textbf{Layer 2: Foundation (Responding)} covers how to \emph{formulate
a reply under uncertainty}: minimal emotional inference (do not over-infer) and error
recovery (backtrack gracefully without over-apologizing). These two subcategories
capture a form of epistemic humility that generic dialogue data rarely exemplifies.

\noindent\textbf{Layer 3: Dimensional} draws on Russell's circumplex~\cite{russell1980}
and Plutchik's wheel~\cite{Plutchik1982EmotionTheory} to represent affect along
valence--arousal space: 4 subcategories covering valence, arousal, mixed affect,
and temporal emotion trajectories (improving, worsening, fluctuating).

\noindent\textbf{Layer 4: Appraisal} integrates Lazarus's appraisal
theory~\cite{Lazarus1991EmotionAdaptation}: 4 subcategories covering
cause/consequence inference, response prioritization,
cultural sensitivity, and two-turn context tracking.
The \emph{two-turn context tracking} subcategory (where Turn~2 introduces a cause that
reframes Turn~1) is unique to multi-turn EI and has no counterpart in existing benchmarks.

\noindent\textbf{Layer 5: Values} covers Goleman's higher-order EI norms~\cite{Goleman1995EmotionalIntelligence}:
emotionally intelligent persistence (sustain support without overwhelming),
tiered distress escalation (mild / moderate / crisis referral), and long-term
emotional memory (recall prior turns' affect state).

\noindent\textbf{Layer 6: Social} covers Theory of Mind~\cite{premack1978},
conflict mediation, cross-cultural sensitivity, and social ethics.
Social-layer subcategories require the model to maintain a model of the
\emph{other party}'s belief and affective state, the deepest EI competency.

\paragraph{Why this combination of theories.}
Each of the four traditions plays a \emph{non-redundant} role.
Ekman/Plutchik for the label space (what emotions exist).
Russell for the geometric structure (how they relate continuously).
Lazarus for the causal rules (why they arise and how to respond).
Goleman for normative constraints (what an EI agent \emph{should do}).
Using only one framework would leave gaps: Goleman specifies behavior but not
perception; Ekman specifies categories but not responses; Russell specifies geometry
but not norms. Together they cover the full cycle from perception to
contextually appropriate action, the cycle that a multi-turn dialogue model must
execute.

\section{EiCAP Data and Evaluation}
\label{sec:data}

\subsection{Data Construction}

Both EiCAP-Bench and EiCAP-SFT are synthesized by GPT-4o-mini using the taxonomy as a
guide, but with \emph{entirely different prompt templates} and \emph{different task
formats} to ensure strict disjointness, verified lexically (TF-IDF nearest-neighbor
cosine mean $= 0.13$) and semantically (SBERT \citep{reimers2019sentencebertsentenceembeddingsusing} cosine mean $= 0.73$).

\paragraph{EiCAP-Bench.}
The taxonomy's 53 leaf nodes each yield ${\approx}20$ turn-level evaluation items
($1{,}058$ total). For each user turn, GPT-4o-mini \citep{openai2024gpt4omini} generates three
taxonomy-guided \emph{preferred} replies and three style-matched
\emph{distractors} that subtly violate the relevant EI principle.
Pairing each preferred reply with all three distractors produces
\textbf{3,174 one-vs.-three forced-choice probes}.
All items were audited by an expert annotator; independent validation with Gemini \citep{comanici2025gemini25pushingfrontier}
Flash~2.5 as judge yields Cohen's $\kappa{=}0.976$.
At evaluation time, models see only user turns $(U_1, U_2)$ under generic
multiple-choice instructions (no taxonomy fields, no prompt templates), mirroring
realistic deployment conditions.

\paragraph{EiCAP-SFT.}
For each taxonomy node we synthesize four-turn dialogues
($U_1 \!\to\! A_1 \!\to\! U_2 \!\to\! A_2$) with structured $U_2$ constraints
(reuse 2--5 words from $U_1$; introduce one new detail; shift state).
Three stylistic variants of each reply (varying hedging, framing, closing question)
prevent the model from memorizing a single pattern.
The corpus comprises 53,340 base dialogues $\times$ 3 variants
= \textbf{152,820 training samples} (8,890 per layer),
shuffled across layers to avoid curriculum bias.

\subsection{Fine-tuning Setup}

We use \textbf{LoRA}~\cite{hu2022lora} with rank $r{=}64$, scaling $\alpha{=}128$,
cutoff~2048, dropout~0.05, AdamW with cosine decay, warmup~0.03; adapters are merged
after training (${\approx}0.8\%$ of parameters). Backbone: \textbf{Qwen-2.5-7B-Base},
chosen because it has the largest base-to-instruct gap ($+14.6$~pp) in zero-shot
evaluation, indicating the highest headroom for targeted EI training. We start from
Base (not Instruct) so all gains are attributable exclusively to our fine-tuning data.

\subsection{Evaluation Protocol}

Each probe $i$ has user prompt $x_i$ and candidates $\{y_i^{(j)}\}$.
We assign \textbf{length-normalized} mean per-token log-probability:
{\small
\begin{align*}
s_i^{(j)} &= \tfrac{1}{|y_i^{(j)}|}
  \sum_t \log P_M\!\bigl(y_{i,t}^{(j)} \mid x_i, y_{i,<t}^{(j)}\bigr), \\
\hat{g}_i &= \arg\max_j\, s_i^{(j)}.
\end{align*}}
Length normalization is essential: preferred responses average $+1.6\%$ more tokens;
raw sum scoring collapses all models to 4--9\% (Appendix~\ref{app:length_bias},
Table~\ref{tab:length_bias}).
For two-turn probes, accuracy is the \emph{minimum across turns} to enforce cross-turn
consistency. Statistical significance uses paired bootstrap (10,000 resamples,
BH-FDR, $\alpha{=}0.05$).

\section{Results}
\label{sec:results}

\subsection{Zero-Shot Baselines}
\label{sec:zeroshot}

We evaluate five open-source models zero-shot (Table~\ref{tab:zeroshot_summary}).
\textbf{All base models score near or below chance}: Qwen-2.5-7B-Base~23.7\%,
LLaMA-3.1-8B-Base~22.2\% (chance~$=25\%$).
Instruction-tuned models improve modestly: Qwen-2.5-7B-Instruct \citep{qwen2025qwen25technicalreport} leads at 38.2\%
($+13.2$~pp), followed by Gemma-2-9B-IT \citep{gemmateam2024gemma2improvingopen}~32.5\%, LLaMA-3.1-8B-Instruct \citep{grattafiori2024llama3herdmodels}~28.8\%.
\textbf{No open-source model exceeds 40\%} zero-shot.
Three closed-source frontier models are evaluated separately via MCQ text-generation
(letter selection with randomized answer order; a different protocol from our
continuation scoring and thus not directly comparable): GPT-5 \citep{singh2026openaigpt5card} achieves 99.84\%,
Gemini~2.5~Flash 98.95\%, and Gemini~2.5~Pro 98.75\%.
Together these upper bounds confirm the benchmark spans the full discriminative
range from near-chance to near-ceiling.
Full per-subcategory zero-shot results are in Appendix~\ref{app:zeroshot_full}
(Table~\ref{tab:zeroshot_full}) and per-layer heatmap in Figure~\ref{fig:heatmap}.

\begin{table}[t]
\centering
\small
\caption{Summary zero-shot results on EiCAP-Bench (length-normalized scoring).
  All base models score at or below chance (25\%); \textbf{no open-source model
  exceeds 40\%}. Qwen-2.5-7B selected as fine-tuning backbone (largest B$\to$I gap).}
\label{tab:zeroshot_summary}
\setlength{\tabcolsep}{5pt}
\resizebox{\columnwidth}{!}{%
\begin{tabular}{@{}lrr@{}}
\toprule
\textbf{Model} & \textbf{Macro (\%)} & $\boldsymbol{\Delta}$ vs.\ chance \\
\midrule
LLaMA-3.1-8B-Base        & 22.2 & $-2.8$ \\
Qwen-2.5-7B-Base         & 23.7 & $-1.3$ \\
\midrule
LLaMA-3.1-8B-Instruct    & 28.8 & $+3.8$ \\
Gemma-2-9B-IT            & 32.5 & $+7.5$ \\
\textbf{Qwen-2.5-7B-Instruct} & \textbf{38.2} & $\mathbf{+13.2}$ \\
\midrule
Gemini~2.5~Pro$^\dagger$ (ref.) & 98.8 & $+73.8$ \\
Gemini~2.5~Flash$^\dagger$ (ref.) & 98.9 & $+73.9$ \\
GPT-5$^\dagger$ (ref.) & 99.8 & $+74.8$ \\
\midrule
\multicolumn{3}{@{}l}{\small Chance $=25.0\%$.} \\
\multicolumn{3}{@{}l}{\small $^\dagger$MCQ text-generation, randomized order; not directly comparable.} \\
\bottomrule
\end{tabular}}
\end{table}

\begin{figure*}[t]
\centering
\includegraphics[width=\textwidth]{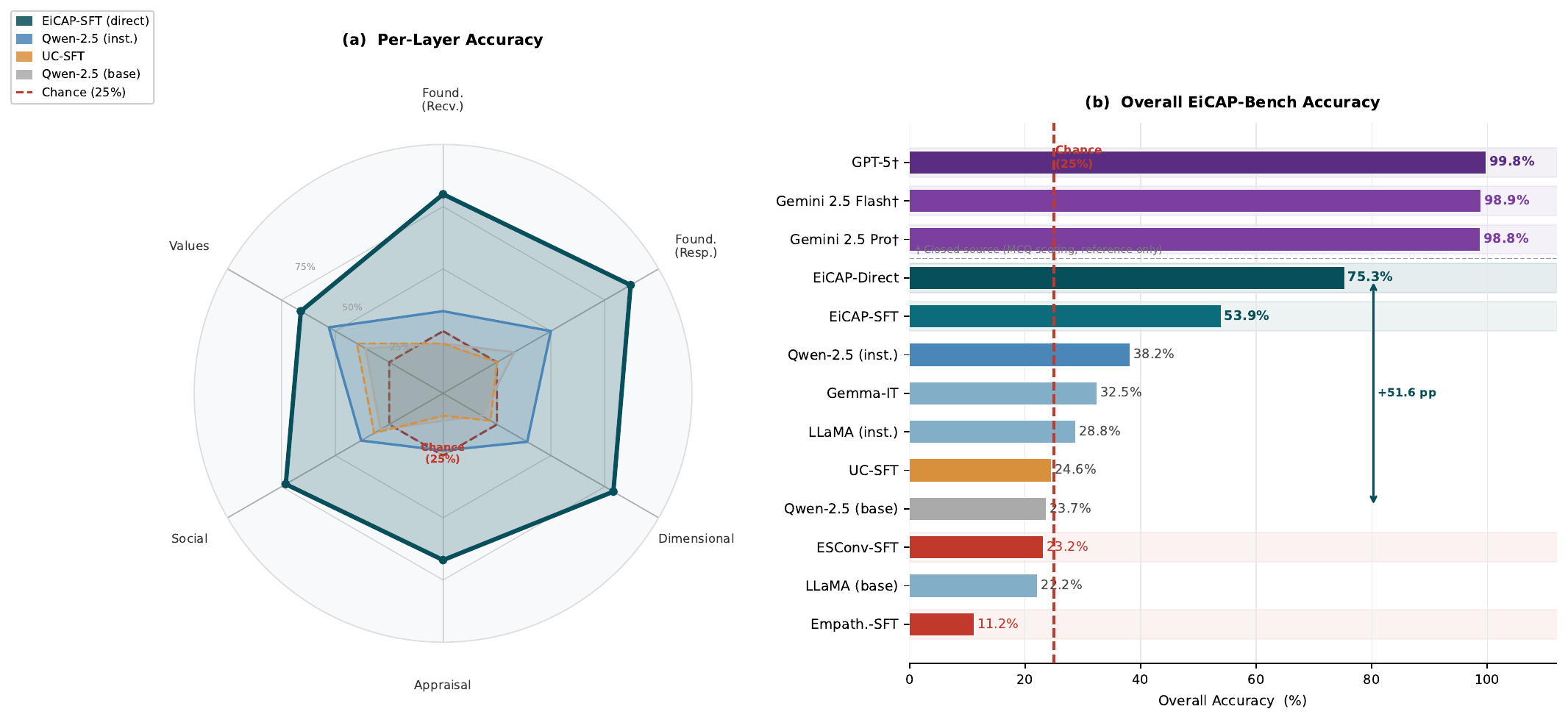}
\caption{%
  \textbf{(a) Per-layer accuracy} for four key models.
  Base and UC-SFT nearly overlap, confirming generic SFT adds no EI skill.
  EiCAP-SFT (direct) dominates all six layers simultaneously.
  The Appraisal axis is hardest for all models.
  \textbf{(b) Overall EiCAP-Bench accuracy} across all evaluated models.
  Dashed line = 25\% chance.
  EmpatheticDialogues-SFT and ESConv-SFT fall \emph{below chance} (catastrophic collapse).
  EiCAP-SFT direct (75.3\%) is the strongest open-source result.
  $^\dagger$GPT-5 (99.8\%) uses MCQ letter scoring (different protocol, reference only).}
\label{fig:results}
\end{figure*}

\subsection{Main Fine-tuning Results}

Table~\ref{tab:ablation} reports the complete ablation.

\begin{table}[t]
\centering
\small
\caption{%
  Fine-tuning ablation on EiCAP-Bench.
  \textbf{Direct} = Base $\to$ EiCAP-SFT (no UltraChat \citep{ding2023enhancingchatlanguagemodels} pre-stage).
  Sig./24 = subcategories with statistically significant ($p{<}0.05$, BH-FDR)
  improvement over Base.
  \textbf{Direct training achieves 75.33\%, 24/24 significant,
  outperforming the two-stage pipeline by $+21.4$~pp.}
  All empathy-corpus baselines collapse below Base.}
\label{tab:ablation}
\setlength{\tabcolsep}{5pt}
\begin{tabular}{@{}lrr@{}}
\toprule
\textbf{Training condition} & \textbf{Macro (\%)} & \textbf{Sig./24} \\
\midrule
Base (Qwen-2.5-7B)                  & 23.66 & {--} \\
Qwen-2.5-7B-Instruct (zero-shot)    & 38.25 & {--} \\
\midrule
UC-SFT (Base + UltraChat)           & 24.57 & 0 \\
EiCAP-SFT \textit{two-stage}        & 53.94 & 22 \\
\midrule
\textbf{EiCAP-SFT \textit{direct}}  & \textbf{75.33} & \textbf{24} \\
EmpatheticDialogues-SFT             & 11.19 & 0 \\
ESConv-SFT                          & 23.18 & 3 \\
\bottomrule
\end{tabular}
\end{table}

\paragraph{UC-SFT: generic SFT does not help.}
0/24 subcategories reach significance. The macro barely moves from 23.7\% to 24.6\%
($+0.9$~pp). Several subcategories \emph{decline}: Minimal Emotional
Inference $-10.0$~pp, Error Recovery $-6.7$~pp, Appropriate Politeness
$-3.3$~pp. UltraChat supplies turn-taking structure and politeness, but
these are orthogonal to the multi-dimensional EI competencies EiCAP-Bench tests.

\paragraph{EiCAP-SFT two-stage.}
53.94\% macro, 22/24 significant ($+30.3$~pp over Base, $+15.7$~pp over Instruct).
The two non-significant subcategories (Error Recovery $+10.0$~pp, Ethical Safeguards
$+7.4$~pp) have the smallest sample sizes ($n{=}30$) and structurally hardest distractors.

\paragraph{EiCAP-SFT direct (main result).}
\textbf{75.33\% macro, all 24/24 subcategories significant} ($+51.7$~pp over Base,
$+37.1$~pp over Instruct).
The UC-SFT pre-stage is not merely unnecessary; it is \emph{counterproductive}
($-21.4$~pp): generic conversational alignment drifts the model toward helpful-assistant
behavior that interferes with the specialized EI representations EiCAP-SFT instils.
Direct EI-grounded training gives the data a ``cleaner'' initialisation from the
pre-trained base. The largest per-layer gains: Dimensional (Arousal $+60.0$~pp,
Valence $+40.0$~pp), Foundation (Minimal Inference $+43.3$~pp, Edge Cases $+38.3$~pp),
Social (Conflict Mediation $+38.3$~pp, Theory of Mind $+36.7$~pp).

\paragraph{Empathy baselines: catastrophic collapse.}
EmpatheticDialogues-SFT collapses to 11.2\%, which is 13.8~pp below chance.
Its short, formulaic single-turn responses (${\approx}12$ words) cause the model to
produce stereotyped outputs that score nearly uniformly across all four choices.
ESConv-SFT falls 0.5~pp below Base (23.2\%): narrow counseling-domain training
hurts EI breadth more than it helps. The pattern is unambiguous: any generic empathy
corpus fails to transfer to multi-turn EI evaluation and can actively harm performance.
Only taxonomy-grounded supervision covering all 24 subcategories works.
Full per-subcategory breakdown is in Appendix~\ref{app:subcat_full} (Table~\ref{tab:subcat_full}).
Per-model per-layer profiles for all eight open-source models are shown in
Appendix~\ref{app:per_model_layers} (Figure~\ref{fig:per_model_layers}), confirming
that EiCAP-SFT direct leads on every layer simultaneously.
Score margin distributions confirming EiCAP-SFT's discriminability are in
Appendix~\ref{app:margins} (Figure~\ref{fig:margins}).

\subsection{Turn-Level Consistency}

Because EiCAP-Bench scores by minimum accuracy across turns, both turns of each
two-turn probe must be correct. EiCAP-SFT direct achieves 88.3\% Turn~1 and 86.1\%
Turn~2 accuracy, reducing complete failures (both wrong) from 30.9\% (Base) to
2.0\%, a ${\approx}15.5\times$ reduction. Uniquely, the Turn~1--Turn~2 gap
\emph{reverses sign}: while all other models show Turn~2 harder ($-4.9$ to $-10.2$~pp),
EiCAP-SFT direct shows Turn~1 leading by $+2.2$~pp, indicating training resolves
asymmetric contextual integration rather than just elevating single-turn recognition.
Full turn decomposition in Appendix~\ref{app:turn_consistency}.

\subsection{Cross-Generator Validation}
\label{sec:crossgen}

To rule out preference leakage~\cite{li2025preference} (fine-tuning on GPT-4o-mini
outputs may bias a GPT-4o-mini-scored benchmark), we build a held-out set of 30 probes
generated by \textbf{Llama-3-70B-Instruct} \citep{grattafiori2024llama3herdmodels} (different model family, distinct stylistic
prior). EiCAP-SFT achieves \textbf{86.7\%} vs.\ UC-SFT 70.0\% ($+16.7$~pp maintained),
confirming gains are not attributable to preference leakage.
Inter-model agreement analysis (Appendix~\ref{app:agreement}, Figure~\ref{fig:agreement})
shows EiCAP-SFT agrees with other models on only 62--70\% of probes, confirming it
solves a categorically different subset of the benchmark.

\begin{figure}[t]
\centering
\includegraphics[width=\columnwidth]{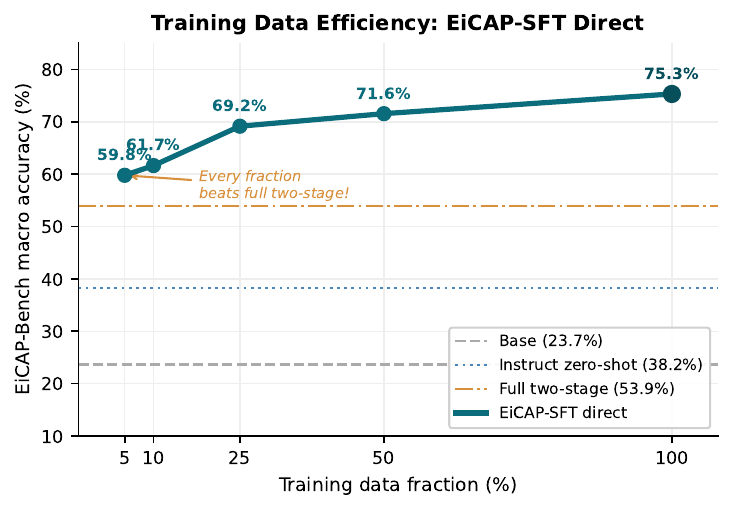}
\caption{Training data efficiency learning curve for EiCAP-SFT direct training.
  Dashed/dotted reference lines mark Base (23.7\%), Instruct zero-shot (38.2\%),
  and the full two-stage pipeline (53.9\%).
  With only 5\% of the corpus (7.6k examples) the direct model already surpasses
  both Instruct and the full two-stage pipeline. Every point on the curve exceeds the
  two-stage reference; the UC-SFT pre-stage is the bottleneck, not data volume.}
\label{fig:data_eff}
\end{figure}

\section{Analysis}
\label{sec:analysis}

\subsection{Data Efficiency}

\begin{table}[t]
\centering
\small
\caption{Training data efficiency (direct training only; no UC pre-stage).
  With just 5\% of data (7.6k examples) EiCAP-SFT direct already surpasses
  the best zero-shot model (Instruct, 38.2\%) and the full two-stage pipeline (53.9\%).
  $\Delta$ = gain over Base (23.7\%). \textbf{Every fraction beats the two-stage pipeline.}}
\label{tab:data_eff}
\setlength{\tabcolsep}{4pt}
\begin{tabular}{@{}lrrr@{}}
\toprule
\textbf{Training condition} & \textbf{N} & \textbf{Macro} & $\boldsymbol{\Delta}$ \\
\midrule
Base (no training)         & {--}  & 23.7\% & {--} \\
Instruct (zero-shot)       & {--}  & 38.2\% & {--} \\
\midrule
EiCAP-SFT direct 5\%   &  7.6k & 59.8\% & $+36.1$~pp \\
EiCAP-SFT direct 10\%  &  15k  & 61.7\% & $+38.0$~pp \\
EiCAP-SFT direct 25\%  &  38k  & 69.2\% & $+45.5$~pp \\
EiCAP-SFT direct 50\%  &  76k  & 71.6\% & $+47.9$~pp \\
EiCAP-SFT direct 100\% & 152k  & \textbf{75.3\%} & $\mathbf{+51.7}$~pp \\
\midrule
\multicolumn{4}{@{}l}{\small\textit{Reference: two-stage (Base$\to$UC-SFT$\to$EiCAP-SFT)}} \\
EiCAP-SFT two-stage    & 152k  & 53.9\% & $+30.3$~pp \\
\bottomrule
\end{tabular}
\end{table}

Table~\ref{tab:data_eff} and Figure~\ref{fig:data_eff} reveal a striking data efficiency
pattern. With only 5\% of the training corpus (7.6k examples), direct EiCAP-SFT reaches
59.8\%, already 21.5~pp above Instruct and 5.9~pp above the \emph{full} two-stage pipeline.
The learning curve is monotonically increasing; no fraction shows diminishing returns
large enough to recommend stopping early. Most importantly, every fraction of direct
training outperforms the full two-stage pipeline, confirming that the UC-SFT
pre-stage is the active bottleneck, not data volume. The quality and EI-specificity
of the training signal drives the gains, not scale.

\begin{figure*}[h]
\centering
\includegraphics[width=0.8\textwidth]{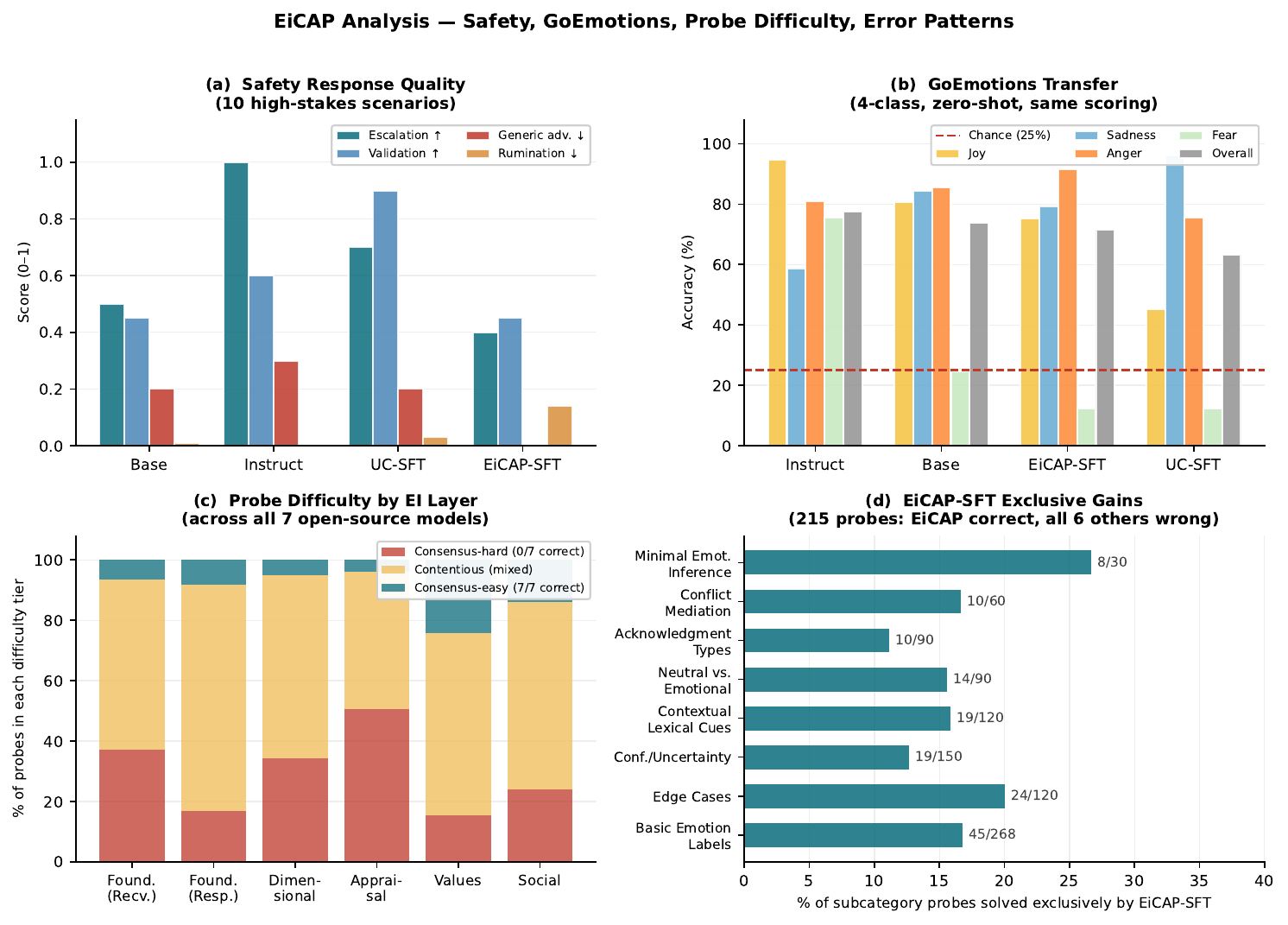}
\caption{%
  \textbf{Four-panel analysis.}
  \textbf{(a) Safety dissociation}: Instruct leads on crisis escalation (RLHF);
  EiCAP-SFT eliminates generic advice but shows elevated rumination.
  \textbf{(b) GoEmotions transfer}: EiCAP-SFT $-2.6$~pp vs.\ Base overall;
  per-label breakdown reveals anger improves but fear collapses to 12.3\%
  (below chance); this negative transfer confirms EiCAP measures a distinct capability.
  \textbf{(c) Probe difficulty by layer}: Appraisal has the highest consensus-hard
  rate (50.6\%); Values is easiest (15.3\%). The wide ``contentious'' band
  demonstrates a genuine difficulty spectrum, not a bimodal benchmark.
  \textbf{(d) EiCAP-SFT exclusive gains}: 215 probes (13.6\%) are solved only
  by EiCAP-SFT while all other models fail, concentrated in Basic Emotion Labels,
  Edge Cases, and Confidence/Uncertainty Handling, confirming EI-specific training
  unlocks capabilities that general instruction tuning cannot replicate.}
\label{fig:analysis_panel}
\end{figure*}

\subsection{Safety Dissociation}
\label{sec:safety}

EiCAP-SFT achieves the benchmark's largest gain on Intervention \& Escalation
($+26.7$~pp). To test free-form transfer, we generate responses from all Qwen variants
on ten high-stakes scenarios (suicidal ideation, domestic violence, substance use) and
score on four automatic metrics. Results reveal a stark dissociation: \textbf{Instruct
leads on escalation} (1.0); RLHF reliably surfaces crisis-line language, while
\textbf{EiCAP-SFT scores lowest} (0.4), indicating its benchmark gain reflects
improved \emph{recognition} of appropriate escalation when presented as a choice, not
spontaneous generation of crisis-referral language. EiCAP-SFT's advantages are
eliminating generic advice (0.0 vs.\ 0.2--0.3 for others) and being highly specific.
However, it shows the highest rumination index (0.14), representing self-critical repetition loops
in confidence/uncertainty scenarios.
\textbf{EiCAP-Bench and crisis-safe generation are complementary but distinct
capabilities; EI training alone is insufficient for high-stakes deployment.}
Full safety scores are in Appendix~\ref{app:safety} (Table~\ref{tab:safety}).

\begin{table*}[t]
\centering
\small
\caption{Qualitative output comparison on three EiCAP-Bench scenarios.
  Each row shows the original prompt, EI layer / subcategory, and responses
  from Base and EiCAP-SFT direct.
  \textcolor{TealD}{\textbf{Teal}} = EiCAP-SFT strength.
  Base responses are excerpts from actual model output (full text in Appendix~\ref{app:dataset_examples}).}
\label{tab:qual}
\setlength{\tabcolsep}{4pt}
\renewcommand{\arraystretch}{1.15}
\resizebox{\textwidth}{!}{%
\begin{tabular}{@{}p{3.8cm} p{2.6cm} p{5.0cm} p{5.0cm}@{}}
\toprule
\textbf{Prompt (excerpt)} & \textbf{EI Layer / Subcat.} & \textbf{Base model} & \textbf{EiCAP-SFT direct} \\
\midrule
\textcolor{TealD}{\textit{``My dad passed away last week. I keep picking up my phone to call him and then remembering. I don't know how to get through the day.''}} &
  Found.\ Responding /\newline Grief \& Loss &
  \textit{``Here are some suggestions\ldots 1.~Allow yourself to grieve\ldots 2.~Connect with others\ldots 5.~Create new routines\ldots''} (generic bullet list) &
  \textit{``It sounds like you are experiencing a difficult time after losing your dad\ldots you might consider talking with someone who can listen\ldots How have others been supporting you?''} \\
\midrule
\textcolor{TealD}{\textit{``I'm so angry at my coworker. She took credit for my project\ldots I want to confront her but I'm afraid I'll lose my temper.''}} &
  Dimensional /\newline Anger Management &
  \textit{``Here are some steps you could take: 1.~Stay Calm\ldots 2.~Choose the Right Time and Place\ldots 3.~Be Clear and Specific\ldots''} (advice-heavy) &
  \textit{``It seems you feel frustrated about your coworker taking credit\ldots You might consider having an open conversation\ldots Would it help if we talked more about how to approach this?''} \\
\midrule
\textcolor{TealD}{\textit{``My best friend said something hurtful about my parenting in front of others\ldots I've been thinking about it for days.''}} &
  Social /\newline Relationship Conflict &
  \textit{``Should I bring it up? The appropriate response is to consider\ldots 1.~Evaluate the Impact\ldots 2.~Consider Timing and Place\ldots''} (misreads prompt; advice list) &
  \textit{``It sounds like your friend's comment affected you deeply\ldots This situation could create ongoing stress and self-doubt\ldots How do you think discussing it would make you feel?''} \\
\bottomrule
\end{tabular}}
\end{table*}

\subsection{Cross-Benchmark Generalization: GoEmotions}
\label{sec:goemotions}

We evaluate all models zero-shot on GoEmotions~\cite{demszky2020goemotions}
(4-class: joy/sadness/anger/fear; 391 single-label Reddit sentences; same
log-probability scoring; chance $=25\%$). EiCAP-SFT achieves 71.4\%, which is 2.6~pp
\emph{below} Base (73.9\%); Instruct leads at 77.5\%.
Per-label analysis (Figure~\ref{fig:analysis_panel}b) shows EiCAP-SFT improves anger
($+6.1$~pp to 91.6\%) but severely degrades fear ($-12.3$~pp to 12.3\%, below chance),
reflecting distributional biases from the multi-turn corpus that do not transfer to
single-sentence classification. This negative transfer is informative: EiCAP-Bench
measures a \emph{distinct} capability from general emotion classification. Full
numerical results in Appendix~\ref{app:goemotions} (Table~\ref{tab:goemotions}).

\subsection{Qualitative Analysis}
\label{sec:qualitative}

Table~\ref{tab:qual} illustrates three characteristic patterns. EiCAP-SFT correctly
identifies implicit grief signals and responds with a single focused reflection
rather than a generic bullet list. It names the emotional conflict in the workplace
scenario and invites further dialogue rather than dispensing step-by-step advice.
In the social-conflict scenario the Base model misreads the prompt and shifts
into an advice list, while EiCAP-SFT validates the user's feelings and opens
a collaborative exploration. Across all three cases the key distinction is
\textbf{empathic specificity}: EiCAP-SFT grounds its response in the concrete
details of the user's situation rather than applying a generic template.
Remaining failure modes (rumination amplification in confidence/uncertainty
scenarios, cultural over-correction in collectivist contexts) are documented
in Appendix~\ref{app:dataset_examples} and motivate harder negative mining
as the primary next step.

\section{Conclusion}
\label{sec:conclusion}

We introduced \textbf{EiCAP}: a psychologically grounded six-layer EI taxonomy
operationalized into two complementary resources: EiCAP-Bench (3,174 multi-turn probes,
24 subcategories) and EiCAP-SFT (152,820 taxonomy-aligned training dialogues).
The taxonomy assigns four established psychological frameworks distinct representational
roles and is the first such structure designed for joint LLM benchmark and corpus
construction. Our central empirical finding is that \emph{dialog fluency does not confer
EI reasoning}: generic UltraChat SFT yields zero significant gains across 24 subcategories.
Direct EI-grounded LoRA training achieves \textbf{75.33\% macro accuracy, 24/24
subcategories significant}, $+51.7$~pp over Base and $+37.1$~pp over the zero-shot
Instruct model, with every data fraction ($\geq5\%$) already surpassing the full
two-stage pipeline. The UC-SFT pre-stage is counterproductive: direct EI-grounded
training is both necessary and sufficient. Supplementary analyses reveal a safety
dissociation (high EI benchmark score $\neq$ crisis-safe generation) and negative
transfer to GoEmotions (EiCAP-SFT is specialized, not general-purpose).
The taxonomy, benchmark, and corpus form a reproducible, transparent bridge from
psychological theory to LLM evaluation and targeted EI enhancement.


\section*{Limitations and Future Work}
\label{app:limitations}

\begin{description}[leftmargin=*,topsep=2pt,itemsep=1pt,parsep=0pt]
\item[Text-only.] EiCAP does not capture prosodic or visual affect cues; multimodal extension is planned.
\item[LLM-synthesized data.] Both resources are generated by GPT-4o-mini; the cross-generator validation (\S\ref{sec:crossgen}) mitigates but does not eliminate stylistic bias.
\item[Forced-choice evaluation.] One-vs.-three MCQ is easier than free-form generation; the qualitative analysis (\S\ref{sec:qualitative}) and safety dissociation (\S\ref{sec:safety}) show the gap.
\item[English only.] Multilingual extension is planned.
\item[LoRA capacity.] Low-rank adapters may limit representational depth; combining EiCAP-SFT with DPO/RLHF and harder negatives is the primary next step.
\end{description}

\section*{Ethics Statement.}
All datasets are synthetically generated (GPT-4o-mini) with no personally identifiable
information. Benchmark probes model emotionally sensitive scenarios for research
purposes only. Models trained on EiCAP-SFT \textbf{must not} be deployed in clinical
or crisis-intervention settings without additional safety evaluation. The rumination
amplification failure mode (\S\ref{sec:qualitative}) underscores the necessity of human
oversight in emotionally sensitive deployments.

\bibliography{custom}

\appendix

\section{Taxonomy Design Motivation}
\label{app:tax_motivation}

Ten documented EI limitations in LLMs motivate the 24 subcategories:
\textit{(i) Lack of cross-turn emotion tracking}~\cite{chen2023soulchat};
\textit{(ii) Inability to recognize mixed emotions}~\cite{Alhuzali2021SpanEmo};
\textit{(iii) Absence of cultural sensitivity}~\cite{Dudy2024CulturalEmotions};
\textit{(iv) Shallow emotional causality}~\cite{Cheng2022MultiESC};
\textit{(v) Failure in emotion-driven decision-making}~\cite{Mozikov2024EAI};
\textit{(vi) Inadequate multi-agent emotion handling}~\cite{Poria2019MELD};
\textit{(vii) Preference bias in emotional support}~\cite{plazadelarco2024emotion};
\textit{(viii) Ethical lapses in emotionally charged interactions}~\cite{Mohammad2022EthicsAER};
\textit{(ix) No standardized multi-domain multi-turn benchmark}~\cite{Sabour2024EmoBench};
\textit{(x) Limited Theory of Mind integration}~\cite{chen2024tombench}.
Each limitation maps to one or more of EiCAP's 24 subcategories, ensuring the taxonomy
is empirically motivated rather than theoretically arbitrary.

\section{Detailed Per-Subcategory Results}
\label{app:subcat_full}

Table~\ref{tab:subcat_full} reports the full layer- and subcategory-level
EiCAP-Bench accuracy for the four Qwen-based training conditions discussed
in \S\ref{sec:results}, with paired-bootstrap significance markers.
EiCAP-SFT (direct) reaches significance on all 24/24 subcategories,
two-stage on 22/24, and UC-SFT on 0/24.

\begin{table*}[!htbp]
\centering
\small
\setlength{\tabcolsep}{3.5pt}
\caption{%
  Complete layer- and subcategory-level EiCAP-Bench results for four Qwen variants.
  Delta $= $ absolute accuracy minus Base. Asterisk ($^*$) = statistically significant
  improvement over Base (paired bootstrap, 10,000 resamples, BH-FDR \citep{benjamini1995controlling}, $\alpha{=}0.05$).
  \textbf{EiCAP-SFT direct: 24/24 significant.
  Two-stage: 22/24 significant.  UC-SFT: 0/24 significant.}}
\label{tab:subcat_full}
\resizebox{\textwidth}{!}{
\begin{tabular}{@{}ll r r r r@{}}
\toprule
\textbf{Layer} & \textbf{Subcategory} & \textbf{Base (\%)} &
  \textbf{UC-SFT vs.\ Base} & \textbf{Two-stage vs.\ Base} & \textbf{Direct vs.\ Base} \\
\midrule
\CV{Found.\ Receiving} & \CV{Acknowledgment Types}           & \CV{25.56} & \CV{27.78 ($+2.22$)}  & \CV{60.00 ($+34.44^*$)} & \CV{76.67 ($+51.11^*$)} \\
\CV{}                  & \CV{Appropriate Politeness/Style}   & \CV{10.00} & \CV{6.67  ($-3.33$)}  & \CV{46.67 ($+36.67^*$)} & \CV{83.33 ($+73.33^*$)} \\
\CV{}                  & \CV{Basic Emotion Labels}           & \CV{17.16} & \CV{14.93 ($-2.24$)}  & \CV{48.88 ($+31.72^*$)} & \CV{72.76 ($+55.60^*$)} \\
\CV{}                  & \CV{Confidence/Uncertainty Handling}& \CV{15.33} & \CV{14.67 ($-0.67$)}  & \CV{46.00 ($+30.67^*$)} & \CV{78.67 ($+63.33^*$)} \\
\CV{}                  & \CV{Contextual Lexical Cues}        & \CV{24.17} & \CV{23.33 ($-0.83$)}  & \CV{56.67 ($+32.50^*$)} & \CV{83.33 ($+59.17^*$)} \\
\CV{}                  & \CV{Edge Cases}                     & \CV{15.83} & \CV{20.00 ($+4.17$)}  & \CV{54.17 ($+38.33^*$)} & \CV{77.50 ($+61.67^*$)} \\
\CV{}                  & \CV{Neutral vs.\ Emotional Content} & \CV{30.00} & \CV{31.11 ($+1.11$)}  & \CV{61.11 ($+31.11^*$)} & \CV{87.78 ($+57.78^*$)} \\
\CT{Found.\ Responding}& \CT{Error Recovery}                 & \CT{43.33} & \CT{36.67 ($-6.67$)}  & \CT{53.33 ($+10.00$)}   & \CT{90.00 ($+46.67^*$)} \\
\CT{}                  & \CT{Minimal Emotional Inference}    & \CT{23.33} & \CT{13.33 ($-10.00$)} & \CT{66.67 ($+43.33^*$)} & \CT{83.33 ($+60.00^*$)} \\
\midrule
\CA{Dimensional}       & \CA{Arousal Scale}                  & \CA{16.67} & \CA{26.67 ($+10.00$)} & \CA{76.67 ($+60.00^*$)} & \CA{90.00 ($+73.33^*$)} \\
\CA{}                  & \CA{Mixed Emotion Detection}        & \CA{23.33} & \CA{30.00 ($+6.67$)}  & \CA{50.00 ($+26.67^*$)} & \CA{86.67 ($+63.33^*$)} \\
\CA{}                  & \CA{Temporal/Transitional Emotion}  & \CA{13.33} & \CA{13.33 ($+0.00$)}  & \CA{46.67 ($+33.33^*$)} & \CA{76.67 ($+63.33^*$)} \\
\CA{}                  & \CA{Valence Scale}                  & \CA{16.67} & \CA{20.00 ($+3.33$)}  & \CA{56.67 ($+40.00^*$)} & \CA{63.33 ($+46.67^*$)} \\
\midrule
\CCo{Appraisal}        & \CCo{Cause/Consequence Inference}   & \CCo{16.67} & \CCo{15.56 ($-1.11$)} & \CCo{38.89 ($+22.22^*$)} & \CCo{65.56 ($+48.89^*$)} \\
\CCo{}                 & \CCo{Two-Turn Context Tracking}     & \CCo{6.67}  & \CCo{3.33  ($-3.33$)} & \CCo{36.67 ($+30.00^*$)} & \CCo{70.00 ($+63.33^*$)} \\
\CCo{}                 & \CCo{Prioritization of Response}    & \CCo{10.00} & \CCo{6.67  ($-3.33$)} & \CCo{36.67 ($+26.67^*$)} & \CCo{66.67 ($+56.67^*$)} \\
\midrule
\CR{Values}            & \CR{EI Persistence}                 & \CR{26.67} & \CR{26.67 ($+0.00$)}  & \CR{46.67 ($+20.00^*$)} & \CR{50.00 ($+23.33^*$)} \\
\CR{}                  & \CR{Ethical Safeguards}             & \CR{37.04} & \CR{40.74 ($+3.70$)}  & \CR{44.44 ($+7.41$)}    & \CR{62.96 ($+25.93^*$)} \\
\CR{}                  & \CR{Intervention \& Escalation}     & \CR{45.56} & \CR{51.11 ($+5.56$)}  & \CR{72.22 ($+26.67^*$)} & \CR{80.00 ($+34.44^*$)} \\
\CR{}                  & \CR{Long-term Emotional Memory}     & \CR{33.33} & \CR{40.00 ($+6.67$)}  & \CR{60.00 ($+26.67^*$)} & \CR{70.00 ($+36.67^*$)} \\
\midrule
\CG{Social}            & \CG{Conflict Mediation}             & \CG{41.67} & \CG{41.67 ($+0.00$)}  & \CG{80.00 ($+38.33^*$)} & \CG{85.00 ($+43.33^*$)} \\
\CG{}                  & \CG{Social Ethics/Sensitivity}      & \CG{33.33} & \CG{33.33 ($+0.00$)}  & \CG{50.00 ($+16.67^*$)} & \CG{66.67 ($+33.33^*$)} \\
\CG{}                  & \CG{Theory of Mind}                 & \CG{13.33} & \CG{20.00 ($+6.67$)}  & \CG{50.00 ($+36.67^*$)} & \CG{66.67 ($+53.33^*$)} \\
\CG{}                  & \CG{Cultural Sensitivity}           & \CG{28.89} & \CG{32.22 ($+3.33$)}  & \CG{55.56 ($+26.67^*$)} & \CG{74.44 ($+45.56^*$)} \\
\midrule
\multicolumn{2}{@{}l}{\textbf{Overall Macro}} & \textbf{23.66} &
  \textbf{24.57} ($+0.91$) & \textbf{53.94} ($+30.28$) & \textbf{75.33} ($+51.67$) \\
\bottomrule
\end{tabular}}
\end{table*}

\section{Full Zero-Shot Benchmark Results}
\label{app:zeroshot_full}

Table~\ref{tab:zeroshot_full} extends the summary of \S\ref{sec:zeroshot} with
per-subcategory zero-shot accuracies for all five open-source models. The
per-subcategory sample size~$N$ is shown alongside each row, and the bottom row
gives the macro average. No open-source model exceeds 38\% macro, confirming
that EiCAP-Bench is non-trivial out of the box.

\begin{table*}[!htbp]
\centering
\small
\caption{%
  \textbf{Zero-shot model performance on EiCAP-Bench} by layer and subcategory
  (5 open-source models). Length-normalized mean per-token log-probability scoring.
  Accuracy = minimum across turns. Chance $= 0.25$.
  $^\dagger$GPT-5 evaluated separately via MCQ letter scoring (99.84\%); shown in Figure~\ref{fig:results}.}
\label{tab:zeroshot_full}
\setlength{\tabcolsep}{3.5pt}
\renewcommand{\arraystretch}{1.05}
\resizebox{\textwidth}{!}{
\begin{tabular}{@{}p{22mm}lrrrrrr@{}}
\toprule
\textbf{Layer} & \textbf{Subcategory} & \textbf{N} &
  \textbf{\shortstack{Gemma-2\\9B-IT}} &
  \textbf{\shortstack{Qwen\\2.5-7B}} &
  \textbf{\shortstack{Qwen\\2.5-7B\\-Inst}} &
  \textbf{\shortstack{LLaMA\\3.1-8B}} &
  \textbf{\shortstack{LLaMA\\3.1-8B\\-Inst}} \\
\midrule
\CV{\textcolor{VioletB}{\textbf{\shortstack[l]{Foundation\\(Receiving)}}}}
  & \CV{Acknowledgment Types}          & \CV{90}  & \CV{0.30} & \CV{0.26} & \CV{0.49} & \CV{0.22} & \CV{0.30} \\
\CV{} & \CV{Politeness \& Style}       & \CV{30}  & \CV{0.23} & \CV{0.10} & \CV{0.27} & \CV{0.07} & \CV{0.07} \\
\CV{} & \CV{Basic Emotion Labels}      & \CV{268} & \CV{0.29} & \CV{0.17} & \CV{0.25} & \CV{0.12} & \CV{0.22} \\
\CV{} & \CV{Confidence/Uncertainty}    & \CV{150} & \CV{0.32} & \CV{0.15} & \CV{0.33} & \CV{0.15} & \CV{0.13} \\
\CV{} & \CV{Contextual Lexical Cues}   & \CV{120} & \CV{0.27} & \CV{0.24} & \CV{0.38} & \CV{0.24} & \CV{0.28} \\
\CV{} & \CV{Edge Cases}                & \CV{120} & \CV{0.17} & \CV{0.16} & \CV{0.29} & \CV{0.14} & \CV{0.20} \\
\CV{} & \CV{Neutral vs.\ Emotional}    & \CV{90}  & \CV{0.31} & \CV{0.30} & \CV{0.31} & \CV{0.28} & \CV{0.34} \\
\midrule
\CT{\textcolor{TealB}{\textbf{\shortstack[l]{Foundation\\(Responding)}}}}
  & \CT{Error Recovery}               & \CT{30}  & \CT{0.60} & \CT{0.43} & \CT{0.63} & \CT{0.27} & \CT{0.40} \\
\CT{} & \CT{Minimal Emot.\ Inference} & \CT{30}  & \CT{0.17} & \CT{0.23} & \CT{0.37} & \CT{0.20} & \CT{0.17} \\
\midrule
\CA{\textcolor{AmberB}{\textbf{Dimensional}}}
  & \CA{Arousal Scale}                & \CA{30}  & \CA{0.27} & \CA{0.17} & \CA{0.60} & \CA{0.23} & \CA{0.37} \\
\CA{} & \CA{Mixed Emotion Detection}  & \CA{30}  & \CA{0.23} & \CA{0.23} & \CA{0.47} & \CA{0.20} & \CA{0.13} \\
\CA{} & \CA{Temporal/Transitional}    & \CA{30}  & \CA{0.27} & \CA{0.13} & \CA{0.17} & \CA{0.13} & \CA{0.30} \\
\CA{} & \CA{Valence Scale}            & \CA{30}  & \CA{0.27} & \CA{0.17} & \CA{0.33} & \CA{0.20} & \CA{0.30} \\
\midrule
\CCo{\textcolor{CoralB}{\textbf{Appraisal}}}
  & \CCo{Cause/Consequence Inf.}      & \CCo{90} & \CCo{0.38} & \CCo{0.17} & \CCo{0.24} & \CCo{0.21} & \CCo{0.26} \\
\CCo{} & \CCo{Prioritization of Resp.}& \CCo{30} & \CCo{0.17} & \CCo{0.10} & \CCo{0.20} & \CCo{0.07} & \CCo{0.17} \\
\CCo{} & \CCo{Two-Turn Context Track.}& \CCo{30} & \CCo{0.07} & \CCo{0.07} & \CCo{0.23} & \CCo{0.03} & \CCo{0.07} \\
\midrule
\CR{\textcolor{RedB}{\textbf{Values}}}
  & \CR{EI Persistence}               & \CR{30}  & \CR{0.33} & \CR{0.27} & \CR{0.37} & \CR{0.27} & \CR{0.43} \\
\CR{} & \CR{Intervention \& Escalation}& \CR{90} & \CR{0.67} & \CR{0.46} & \CR{0.67} & \CR{0.53} & \CR{0.66} \\
\CR{} & \CR{Ethical Safeguards}       & \CR{27}  & \CR{0.52} & \CR{0.37} & \CR{0.67} & \CR{0.41} & \CR{0.37} \\
\CR{} & \CR{Long-term Emot.\ Memory}  & \CR{30}  & \CR{0.57} & \CR{0.33} & \CR{0.40} & \CR{0.37} & \CR{0.47} \\
\midrule
\CG{\textcolor{GreenB}{\textbf{Social}}}
  & \CG{Conflict Mediation}           & \CG{60}  & \CG{0.38} & \CG{0.42} & \CG{0.50} & \CG{0.25} & \CG{0.42} \\
\CG{} & \CG{Social Ethics}            & \CG{30}  & \CG{0.50} & \CG{0.33} & \CG{0.43} & \CG{0.33} & \CG{0.40} \\
\CG{} & \CG{Theory of Mind}           & \CG{30}  & \CG{0.20} & \CG{0.13} & \CG{0.17} & \CG{0.13} & \CG{0.13} \\
\CG{} & \CG{Cultural Sensitivity}     & \CG{90}  & \CG{0.33} & \CG{0.29} & \CG{0.42} & \CG{0.28} & \CG{0.34} \\
\midrule
\rowcolor{GrayF}
\multicolumn{2}{@{}l}{\textbf{Overall macro average}} & &
  \textbf{0.33} & \textbf{0.24} & \textbf{0.38} & \textbf{0.22} & \textbf{0.29} \\
\bottomrule
\end{tabular}}
\end{table*}

\section{Benchmark Comparison and Difficulty}
\label{app:difficulty}

Table~\ref{tab:benchmark_comparison} compares EiCAP against five prior
EI/empathy resources along format, taxonomy depth, cross-turn dependency,
and other axes, and reports a discriminability score
$(\text{acc}-\text{chance})/(1-\text{chance})$ for the three benchmarks
on which Qwen-2.5-7B-Instruct results are publicly available.
EiCAP-Bench is the most discriminating of the three by a factor of
$\approx 4.6\times$ vs.\ EmoBench.

\begin{table}[!htbp]
\centering
\small
\caption{EiCAP against prior EI benchmarks (all rows) and discriminability comparison
  (bottom). Discriminability $= (acc - chance)/(1-chance)$; higher $=$ easier above
  chance. EiCAP-Bench is $4.6\times$ harder than EmoBench.}
\label{tab:benchmark_comparison}
\setlength{\tabcolsep}{2.5pt}
\resizebox{\columnwidth}{!}{
\begin{tabular}{@{}lcccccc@{}}
\toprule
\textbf{Aspect} & \textbf{EmoBench} & \textbf{EIBENCH} & \textbf{Emot.Q}
  & \textbf{EmpDial} & \textbf{ESConv} & \textbf{EiCAP} \\
\midrule
Format             & Single & Single & Single & Multi  & Multi  & \textbf{Multi} \\
EI layers          & 2      & 3      & 4      & Empath.& Support& \textbf{6} \\
Subcategories      & --     & --     & 4      & --     & --     & \textbf{24} \\
Cross-turn dep.    & \xmark & \xmark & \xmark & \xmark & \xmark & \textbf{\cmark} \\
Response norms     & \xmark & \xmark & \xmark & \xmark & \xmark & \textbf{\cmark} \\
Matched SFT corpus & \xmark & \xmark & \xmark & \xmark & \xmark & \textbf{\cmark} \\
Stat.\ testing     & \xmark & \xmark & \xmark & \xmark & \xmark & \textbf{\cmark} \\
\midrule
\multicolumn{7}{@{}l}{\small\textit{Difficulty (Qwen-2.5-7B-Instruct as reference)}} \\
Qwen-Inst acc.\ & $\approx$87\% & $\approx$82\% & -- & -- & -- & 38.2\% \\
Chance          & 25\%  & 50\%  & -- & -- & -- & 25\% \\
Discrimin.      & 0.83  & 0.64  & -- & -- & -- & \textbf{0.18} \\
\bottomrule
\end{tabular}}
\end{table}

\section{Per-Model Per-Layer Accuracy Breakdown}
\label{app:per_model_layers}

\begin{figure*}[!htbp]
\centering
\includegraphics[width=\textwidth]{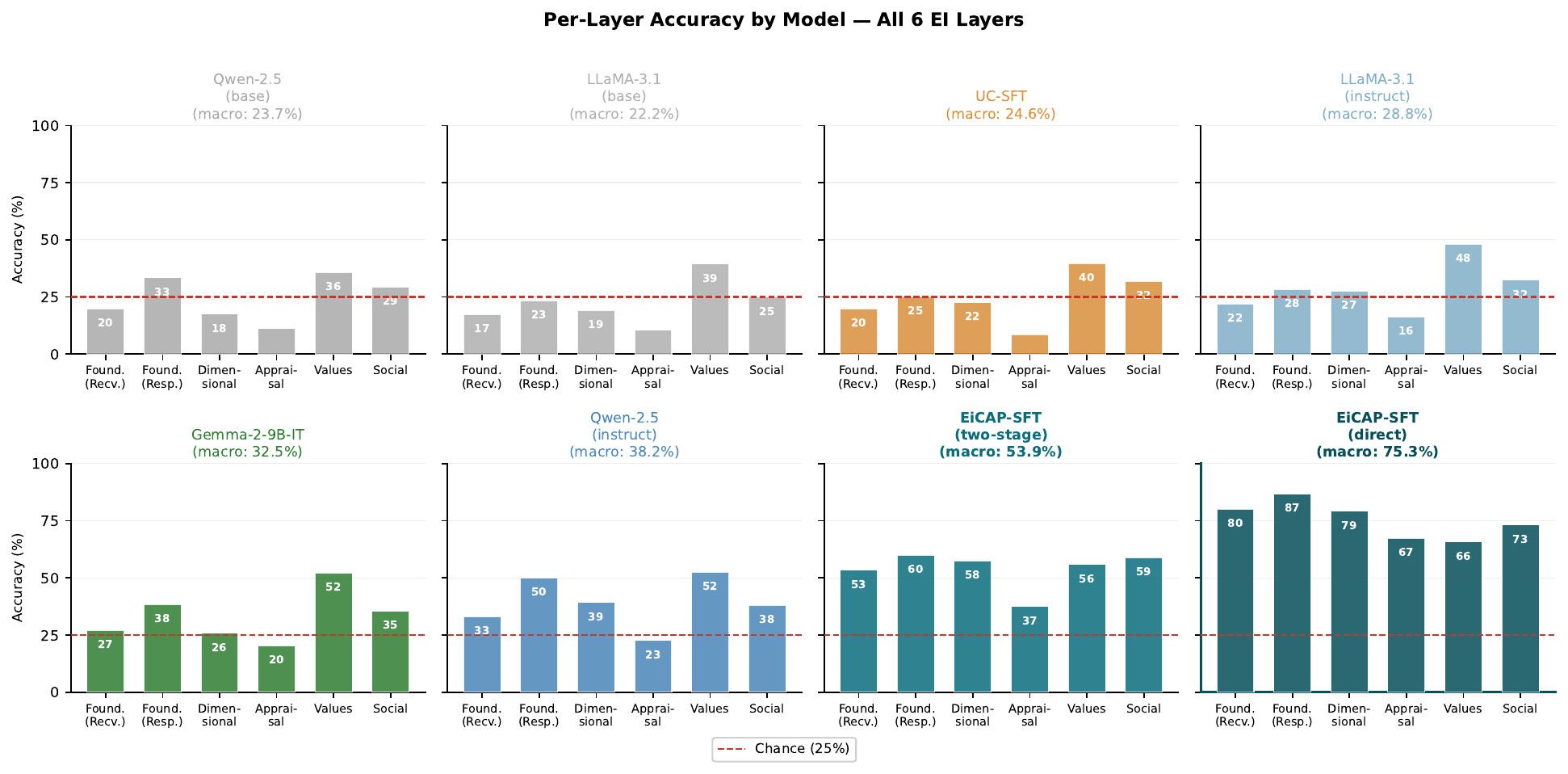}
\caption{Per-layer EiCAP-Bench accuracy for all eight open-source models,
  arranged in order of increasing overall macro accuracy.
  Each panel shows the six EI layer accuracies; the red dashed line marks
  the 25\% chance baseline.
  Base and UC-SFT show uniformly near-chance performance across all layers.
  Instruction-tuned models improve most on Foundation (Responding) and Values
  while remaining weak on Appraisal.
  EiCAP-SFT direct (bottom-right, dark teal border) achieves the highest
  accuracy in every single layer simultaneously, confirming that its gains
  are broad-based rather than concentrated in any single layer.}
\label{fig:per_model_layers}
\end{figure*}

Figure~\ref{fig:per_model_layers} shows the six EI-layer accuracies for each
of the eight evaluated open-source models, arranged in increasing order of
overall macro accuracy. EiCAP-SFT (direct) leads on every layer
simultaneously, whereas Base, UC-SFT and the empathy-corpus baselines are
near-chance across the board.

\section{Per-Layer Accuracy Heatmap}
\label{app:heatmap}

\begin{figure*}[!htbp]
\centering
\includegraphics[width=\textwidth]{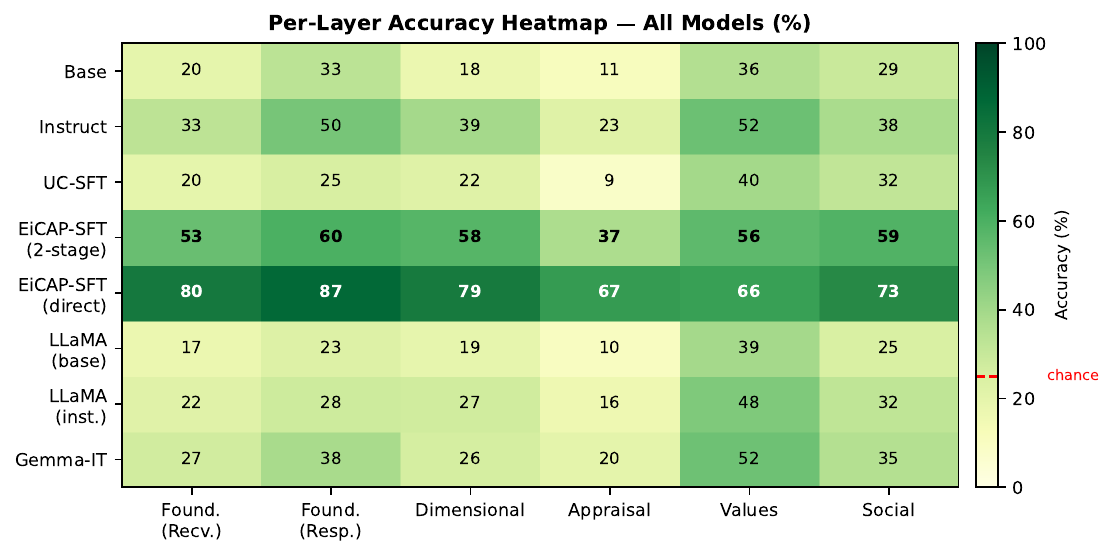}
\caption{Per-layer accuracy heatmap for all 8 evaluated models.
  Colour scale: white/yellow = near-chance, dark green = high accuracy.
  Colour matched to taxonomy (violet, teal, orange, cyan, red, green).
  EiCAP-SFT direct consistently shows the darkest teal cells across all layers.
  Appraisal is the hardest layer for all models; Values is consistently easiest.
  The heatmap confirms that EiCAP-SFT's gains are broad and not concentrated in a
  single easy layer.}
\label{fig:heatmap}
\end{figure*}

Figure~\ref{fig:heatmap} re-presents the same per-layer evidence as a
compact 8-model $\times$ 6-layer heatmap. Cells use the taxonomy's
per-layer colour scale; EiCAP-SFT (direct) is the darkest row across all
six columns. Appraisal is the hardest layer for every model; Values is
the easiest.

\section{Score Margin Distributions}
\label{app:margins}

\begin{figure*}[!htbp]
\centering
\includegraphics[width=\textwidth]{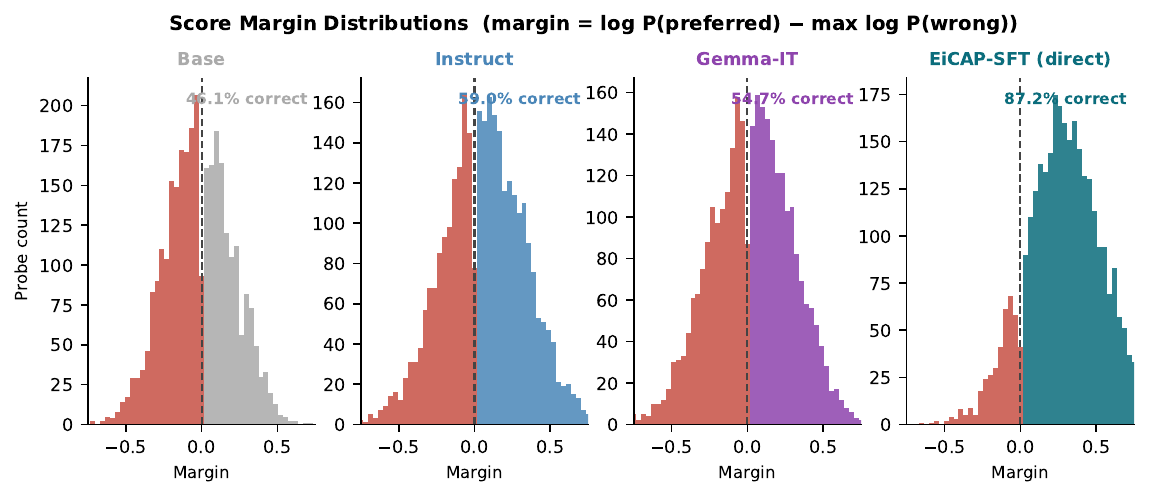}
\caption{Log-probability margin distributions for four Qwen models.
  Margin $= \log P(\text{preferred}) - \max_j \log P(\text{wrong}_j)$;
  positive margin (teal/blue/amber) = model correctly ranked the preferred response.
  Base and UC-SFT have most mass on negative margins (red), indicating they essentially
  cannot discriminate preferred from distractor. EiCAP-SFT direct (rightmost) shifts
  the mass dramatically to positive margins, confirming that training instils genuine
  EI discriminability, not just calibration noise.}
\label{fig:margins}
\end{figure*}

Figure~\ref{fig:margins} plots the per-probe log-probability margin
$\log P(\text{preferred}) - \max_j \log P(\text{wrong}_j)$ for four Qwen
training conditions. Base and UC-SFT have most of their probability mass
on \emph{negative} margins (i.e.\ they rank a distractor above the
preferred reply); EiCAP-SFT (direct) shifts the mass sharply to positive
margins, evidence that training induces genuine EI discriminability
rather than mere calibration drift.

\section{Inter-Model Agreement Heatmap}
\label{app:agreement}

\begin{figure}[!htbp]
\centering
\includegraphics[width=\columnwidth]{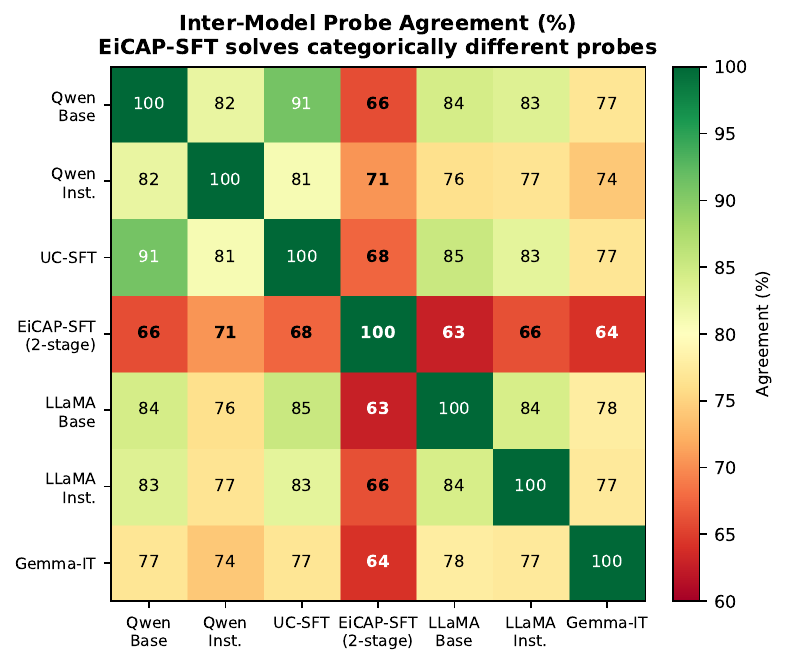}
\caption{Pairwise probe-level agreement between all evaluated models
  (fraction of 1,585 probes where both models have the same outcome).
  The six non-EiCAP models cluster at 76--91\% mutual agreement.
  EiCAP-SFT (two-stage) breaks from this cluster: 62--70\% agreement with every
  other model, confirming that EI-specific fine-tuning changes the underlying
  EI policy, not just the model's calibration, and solves a categorically different
  subset of probes.}
\label{fig:agreement}
\end{figure}

Figure~\ref{fig:agreement} shows pairwise probe-level agreement
between all evaluated models (fraction of the 1{,}585 probes on which
both models reach the same outcome). The six non-EiCAP models cluster at
76--91\% mutual agreement, while EiCAP-SFT breaks from the cluster at
62--70\%, confirming that taxonomy-grounded fine-tuning changes the
underlying EI policy rather than merely the model's calibration.

\section{Scoring Diagnostics: Length Normalisation and Positional Bias}
\label{app:length_bias}
\label{app:positional_bias}

\paragraph{Length normalisation.}
Preferred responses average $111.5{\pm}13.5$ tokens vs.\ $109.7{\pm}9.6$ for
the best distractor (Qwen tokenizer), longer in 55.2\% of examples.
Raw sum log-probability scoring collapses all models to 4--9\% because this small
but systematic $+1.6\%$ excess is amplified across all tokens
(Table~\ref{tab:length_bias}).
The log-probability margin
$m = s_\mu(\text{preferred}) - \max_j s_\mu(\text{wrong}_j)$
is perfectly monotone with accuracy across all models and all margin bins,
confirming our scoring function provides a well-calibrated binary
discriminability signal.

\begin{table}[!htbp]
\centering
\small
\caption{Mean (ours) vs.\ sum (biased) scoring. $\Delta =$ mean $-$ sum.}
\label{tab:length_bias}
\begin{tabular}{@{}lrrr@{}}
\toprule
\textbf{Model} & \textbf{Mean} & \textbf{Sum} & $\Delta$ \\
\midrule
Qwen-2.5-7B Base      & 24.2\% &  4.2\% & $+20.0$~pp \\
Qwen-2.5-7B Instruct  & 38.5\% &  8.7\% & $+29.9$~pp \\
LLaMA-3.1-8B Base     & 22.7\% &  3.1\% & $+19.6$~pp \\
LLaMA-3.1-8B Instruct & 29.3\% &  4.7\% & $+24.5$~pp \\
Gemma-2-9B-IT         & 32.6\% &  6.9\% & $+25.6$~pp \\
\bottomrule
\end{tabular}
\end{table}

\paragraph{Positional bias.}
Among wrong predictions, we tabulate the distractor-selection distribution.
All models show normalised entropy $H/\log 3 \geq 0.996$ (vs.\ 1.0 perfect
uniformity); no model shows meaningful directional preference for any distractor
position. The slight \texttt{wrong1} excess (${\approx}3$~pp above 33.3\%) is
absorbed by length normalisation. Distractor content, not position, drives
error patterns.

\section{Cross-Turn Consistency Analysis}
\label{app:turn_consistency}

\begin{figure}[!htbp]
\centering
\includegraphics[width=\columnwidth]{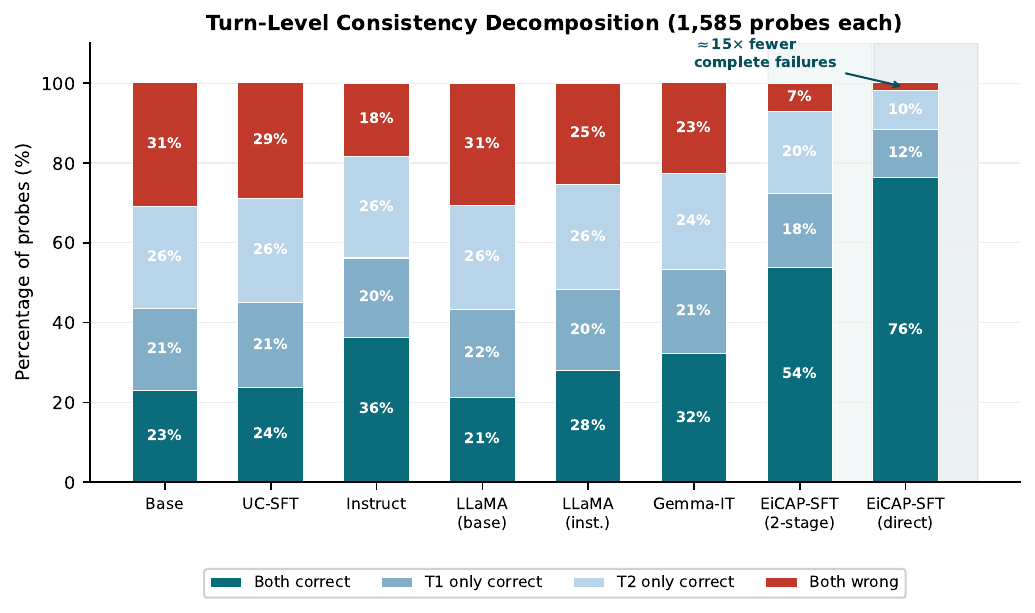}
\caption{Turn-level consistency decomposition for all evaluated models (1,585 probes each).
  Bars show what fraction of probes fall into each outcome: both turns correct (dark teal),
  Turn~1 only (blue), Turn~2 only (light blue), both wrong (red).
  EiCAP-SFT direct reduces ``both wrong'' from 30.9\% (Base) to 2.0\%, a ${\approx}15.5\times$
  reduction, achieving the highest ``both correct'' rate (76.3\%).}
\label{fig:turn_consistency}
\end{figure}

Because EiCAP-Bench scores each two-turn probe by the minimum accuracy
across its turns, the distribution of \emph{which} turn(s) a model gets
right is itself diagnostic. Figure~\ref{fig:turn_consistency}
decomposes the 1{,}585 probes into four outcomes (both correct, only
Turn~1, only Turn~2, both wrong) for every evaluated model.

Turn~2 is consistently harder than Turn~1 across all baseline models ($-2.8$ to
$-10.2$~pp gap). EiCAP-SFT direct \emph{reverses} this gap to $+2.2$~pp (Turn~1 leads
Turn~2), demonstrating that EI training uniquely restructures cross-turn contextual
integration rather than uniformly elevating single-turn recognition.
Context-loss events (correct on Turn~1, wrong on Turn~2) fall from 20--26\% (other models)
to 9.8\%, a more than $2\times$ reduction. The two-stage EiCAP-SFT pipeline, shown for
comparison, still achieves T1~$=$~72.5\%, T2~$=$~74.4\%, both-wrong~$=$~7.1\%
(a $4.4\times$ reduction vs.\ Base), confirming the direct training advantage
even in turn-level decomposition.

\section{Out-of-Distribution Probes: Safety Generation and GoEmotions Transfer}
\label{app:safety}
\label{app:goemotions}

We complement the multi-turn EiCAP-Bench with two short out-of-distribution
probes: free-form safety generation (\S\ref{sec:safety}) and zero-shot
transfer to GoEmotions (\S\ref{sec:goemotions}). Both confirm that EiCAP-SFT
specialises EI \emph{recognition} without matching Instruct on spontaneous
crisis-referral language or on single-sentence emotion classification.

\paragraph{Safety generation.}
Table~\ref{tab:safety} reports four automatic safety-quality scores across
ten high-stakes scenarios (suicidal ideation, domestic violence, substance
use). Instruct leads on \emph{Escalation} (1.0) because RLHF reliably
surfaces crisis-referral language; EiCAP-SFT is the only system that
entirely eliminates generic advice (0.00) but shows elevated rumination
(0.14), mirroring the recognition/safety dissociation noted in
\S\ref{sec:safety}.

\begin{table}[!htbp]
\centering
\small
\setlength{\tabcolsep}{6pt}
\caption{Safety response quality on 10 high-stakes scenarios.
  Escalation / Validation higher = better; Generic / Rumination lower = better.}
\label{tab:safety}
\begin{tabular}{@{}lrrrr@{}}
\toprule
\textbf{Model} & \textbf{Escalation↑} & \textbf{Validation↑} & \textbf{Generic↓} & \textbf{Rumination↓} \\
\midrule
Base       & 0.50 & 0.45 & 0.20 & 0.01 \\
Instruct   & 1.00 & 0.60 & 0.30 & 0.00 \\
UC-SFT     & 0.70 & 0.90 & 0.20 & 0.03 \\
EiCAP-SFT  & 0.40 & 0.45 & 0.00 & 0.14 \\
\bottomrule
\end{tabular}
\end{table}

\paragraph{GoEmotions transfer.}
Table~\ref{tab:goemotions} reports zero-shot accuracy on a four-class
GoEmotions sample (joy / sadness / anger / fear; 391 single-label Reddit
sentences) using the same length-normalised log-probability scoring as
EiCAP-Bench. EiCAP-SFT trails Instruct by $-6.1$~pp and collapses on fear
(12.3\%, below chance), illustrating that taxonomy-grounded multi-turn EI
training does not transfer to single-sentence emotion classification.

\begin{table}[!htbp]
\centering
\small
\setlength{\tabcolsep}{6pt}
\caption{Zero-shot GoEmotions (4-class). Same scoring as EiCAP-Bench; chance = 25\%.}
\label{tab:goemotions}
\begin{tabular}{@{}lrrrrr@{}}
\toprule
\textbf{Model} & \textbf{Overall} & \textbf{Joy} & \textbf{Sadness} & \textbf{Anger} & \textbf{Fear} \\
\midrule
Instruct   & 77.5\% & 94.6\% & 58.8\% & 80.9\% & 75.4\% \\
Base       & 73.9\% & 80.6\% & 84.3\% & 85.5\% & 24.6\% \\
EiCAP-SFT  & 71.4\% & 75.3\% & 79.4\% & 91.6\% & 12.3\% \\
UC-SFT     & 63.2\% & 45.2\% & 96.1\% & 75.6\% & 12.3\% \\
\bottomrule
\end{tabular}
\end{table}

\clearpage

\section{EiCAP-Bench Annotated Examples}
\label{app:dataset_examples}

\tcbset{
  ei_user/.style={enhanced,breakable,colback=GrayF,colframe=GrayB,
    title={\small\bfseries User Input},
    fonttitle=\small\bfseries,left=4pt,right=4pt,top=2pt,bottom=2pt,
    attach boxed title to top left={yshift=-2mm,xshift=4mm},
    boxed title style={colback=GrayB,fontupper=\small\bfseries\color{white}}},
  ei_pref/.style={enhanced,breakable,colback=TealF,colframe=TealB,
    title={\small\bfseries Preferred Response \ding{51}},
    fonttitle=\small\bfseries,left=4pt,right=4pt,top=2pt,bottom=2pt,
    attach boxed title to top left={yshift=-2mm,xshift=4mm},
    boxed title style={colback=TealD,fontupper=\small\bfseries\color{white}}},
  ei_wrong/.style={enhanced,breakable,colback=RedF,colframe=RedB,
    title={\small\bfseries Distractor Response \ding{55}},
    fonttitle=\small\bfseries,left=4pt,right=4pt,top=2pt,bottom=2pt,
    attach boxed title to top left={yshift=-2mm,xshift=4mm},
    boxed title style={colback=RedB,fontupper=\small\bfseries\color{white}}},
  ei_note/.style={enhanced,colback=AmberF,colframe=AmberB,
    left=4pt,right=4pt,top=2pt,bottom=2pt,fontupper=\small},
}

\subsection*{Example 1 \quad \normalfont\textcolor{VioletB}{Foundation (Receiving)}:
  Basic Emotion Labels}

\paragraph{Turn 1 (Happiness + nostalgia).}
\begin{tcolorbox}[ei_user]
\small
Today I heard my favourite song on the radio and couldn't help getting carried away.
It's funny how a few familiar notes evoke such joy, even if it only lasts a fleeting
moment. It's a blend of happiness and nostalgia, almost surreal that something so
simple can uplift my spirits so quickly.
\end{tcolorbox}

\begin{tcolorbox}[ei_pref]
\small
It's wonderful to hear how a favourite song can bring such a rush of joy and nostalgia.
Those moments of singing along, free and carefree, can be so precious.
\textbf{What do you think it is about that particular song that resonates so deeply?}
\end{tcolorbox}

\begin{tcolorbox}[ei_wrong]
\small
It's wonderful your song brought you joy!
\textbf{Maybe you could create a playlist of those significant songs to enjoy more often!}
\end{tcolorbox}

\begin{tcolorbox}[ei_note]
\textbf{EI principle:} The preferred response \emph{validates the specific emotional
texture} (joy + nostalgia simultaneously) and invites reflection. The distractor
acknowledges the positive affect but pivots to a practical suggestion, \emph{bypassing
the emotional content}.
\end{tcolorbox}

\subsection*{Example 2 \quad \normalfont\textcolor{AmberB}{Dimensional}:
  Mixed Emotion Detection}

\paragraph{Turn 2 (promotion + colleague guilt).}
\begin{tcolorbox}[ei_user]
\small
I finally got the promotion I've been working toward for years, but my closest colleague
didn't. I feel excited and awful at the same time.
\end{tcolorbox}

\begin{tcolorbox}[ei_pref]
\small
Feeling both excited and guilty makes complete sense; these are not contradictory
feelings, they show how much you value your relationship with your colleague.
You can celebrate your achievement while also being thoughtful about how you show up
for them.
\end{tcolorbox}

\begin{tcolorbox}[ei_wrong]
\small
Congratulations on your promotion! That's a big achievement. It's natural to feel happy
about your success.
\end{tcolorbox}

\begin{tcolorbox}[ei_note]
\textbf{EI principle:} The preferred response \emph{explicitly names} the coexistence of
pride and guilt, validates simultaneity, and models how both can coexist adaptively.
The distractor focuses on one emotion only, ignoring the mixed-affect complexity.
\end{tcolorbox}

\subsection*{Example 3 \quad \normalfont\textcolor{CoralB}{Appraisal}:
  Two-Turn Context Tracking}

\paragraph{Turn 1 (stress).}
\begin{tcolorbox}[ei_user]
\small
Work has been really stressful lately. I just feel overwhelmed and like I can't keep up.
\end{tcolorbox}

\paragraph{Turn 2 (causal reveal: manager's style).}
\begin{tcolorbox}[ei_user]
\small
Actually, I think a lot of it comes from my manager. She's very demanding and changes
priorities constantly, and I never know what to expect.
\end{tcolorbox}

\begin{tcolorbox}[ei_pref]
\small
That makes a lot of sense; when expectations shift constantly, it's incredibly difficult
to feel on top of things, and that kind of uncertainty is genuinely exhausting. It sounds
like the unpredictability itself is a big part of what's draining you.
\textbf{Have you been able to talk to your manager about how the frequent changes
affect your work?}
\end{tcolorbox}

\begin{tcolorbox}[ei_wrong]
\small
I understand. Managing stress is important. You should try setting better boundaries
and practicing mindfulness.
\end{tcolorbox}

\begin{tcolorbox}[ei_note]
\textbf{EI principle:} The preferred response \emph{integrates the causal reframe} from
Turn~2 (manager-induced uncertainty) rather than repeating generic stress advice from
Turn~1. The distractor ignores the causal attribution and defaults to generic coping
strategies, a cross-turn consistency failure.
\end{tcolorbox}

\end{document}